\PassOptionsToPackage{sort&compress}{natbib}
\documentclass[manuscript,screen,nonacm]{acmart}
\AtBeginDocument{}
\usepackage{amsthm}
\usepackage{booktabs}
\usepackage{cleveref}
\usepackage{tikz}
\usetikzlibrary{positioning,arrows.meta,fit,backgrounds,calc}
\definecolor{fastblue}{HTML}{1F4E79}
\definecolor{slowteal}{HTML}{2A9D8F}

\newcommand{\nrec}{n_{\mathrm{rec}}}

\newtheorem{theorem}{Theorem}[section]
\newtheorem{lemma}[theorem]{Lemma}
\newtheorem{proposition}[theorem]{Proposition}

\AddToHook{env/lemma/begin}{\crefalias{theorem}{lemma}}
\AddToHook{env/proposition/begin}{\crefalias{theorem}{proposition}}
\AddToHook{env/corollary/begin}{\crefalias{theorem}{corollary}}
\graphicspath{{figures/}}
\makeatletter
\let\@ACM@checkaffil\relax
\makeatother
\setcopyright{none}
\begin{document}
\title[Aligning, Fast and Slow]{Aligning, Fast and Slow: A Control-Theoretic Oversight Framework for Long-Horizon Trajectories and Maintained Deployments}
\author{Stephen Peng}
\authornote{Equal contribution.}
\affiliation{\institution{BCG X AI Science Institute, Boston Consulting Group}}
\author{Vineet Jagadeesan Nair}
\authornotemark[1]
\affiliation{\institution{BCG X AI Science Institute, Boston Consulting Group}}
\author{Noah Broestl}
\affiliation{\institution{BCG X AI Science Institute, Boston Consulting Group}}
\affiliation{\institution{Centre for Human-Inspired Artificial Intelligence, University of Cambridge}}
\author{Anne Friederike Kleppe}
\affiliation{\institution{BCG X AI Science Institute, Boston Consulting Group}}
\author{Steven Mills}
\affiliation{\institution{BCG X AI Science Institute, Boston Consulting Group}}
\author{Adi Zolotov}
\affiliation{\institution{BCG X AI Science Institute, Boston Consulting Group}}
\renewcommand{\shortauthors}{Peng et al.}
\authorsaddresses{}

\begin{abstract}
AI agents can pass checks on individual tasks while retaining memory or code changes that alter their later behavior.
We introduce a control-theoretic framework that couples a fast loop controlling proposed actions with a slow loop reviewing persistent changes and reassessing configurations after model or tool updates.
When measurement errors and changes between checks are bounded and controllers can enforce their decisions, we derive a deployment-level bound and a rule for dividing review effort.
In simulation, dividing a fixed review budget between both loops reduces mean maximum departure from intended behavior roughly twelvefold compared with review confined to individual runs.
In coding stress tests with authored memories and five matched seeds, combined controls reduce mean severity on a 0--3 scale from 0.325 without gates to 0.117, resolving 33 of 120 issues versus 35 without gates; in trading stress tests, combined controls reduce mean shortfall in low-emission purchases from 0.892 without gates to 0.107.
The overall framework helps developers and auditors compare and enforce oversight designs, while the agent studies test component controls rather than the full deployment-level guarantee.
\end{abstract}

\maketitle
\hypersetup{
  pdftitle={Aligning, Fast and Slow: A Control-Theoretic Oversight Framework for Long-Horizon Trajectories and Maintained Deployments},
  pdfauthor={Stephen Peng, Vineet Jagadeesan Nair, Noah Broestl, Anne Friederike Kleppe, Steven Mills, Adi Zolotov}
}

\begin{figure}[t]
  \centering
  \includegraphics[width=1.0\textwidth]{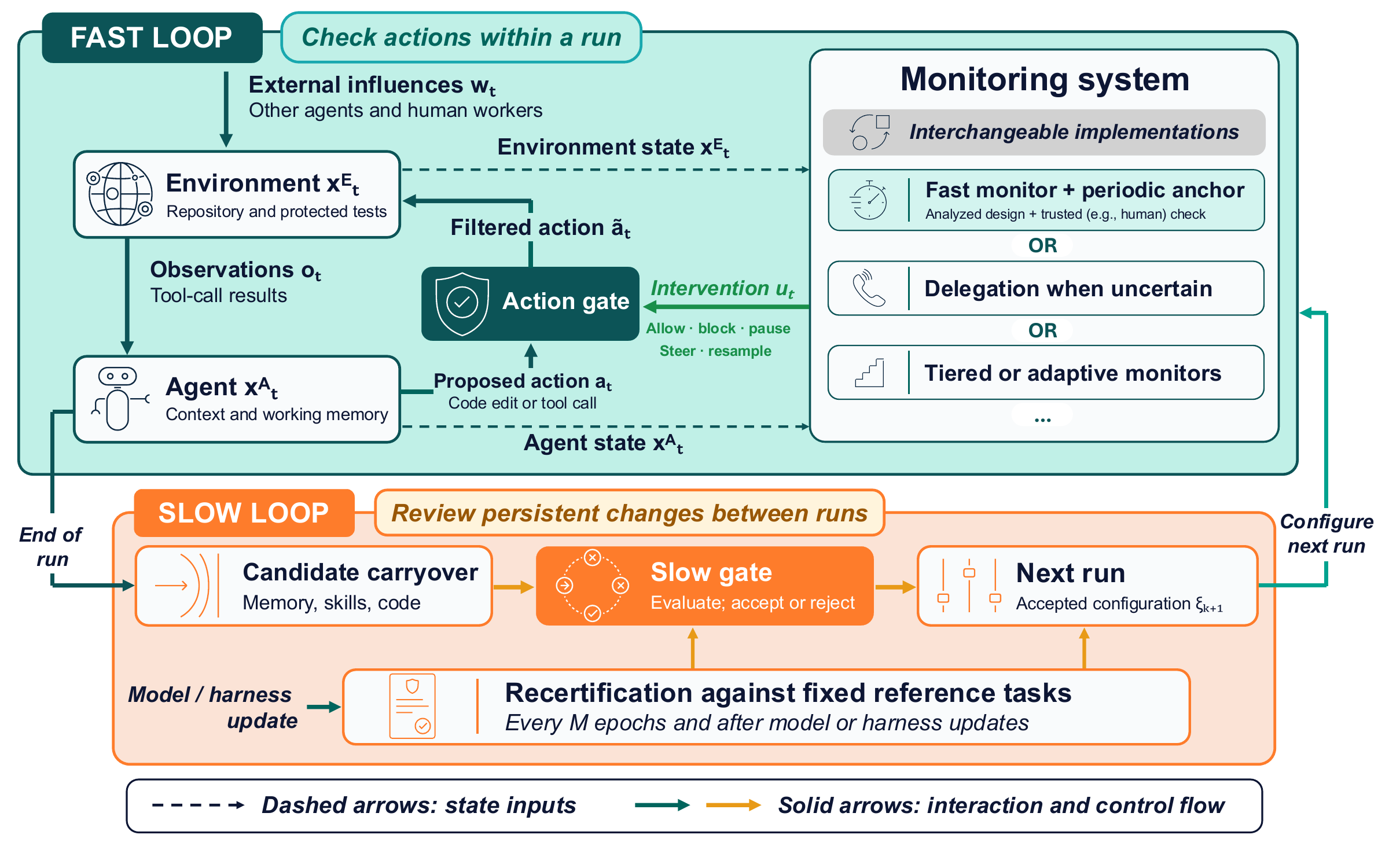}
  \caption{Oversight must cover both an agent's actions and what it leaves behind for the next run. The fast loop checks proposed actions within a run; the slow loop checks persistent changes before later runs use them. A monitor that judges only the current run may be unaware or unable to evaluate inherited memory.}
  \Description{Two connected loops. The fast loop connects the environment, agent, action gate, and monitor. Separate monitor designs are shown as alternatives. The slow loop checks persistent updates and recertifies the configuration, including after external updates.}
  \label{fig:teaser}
\end{figure}

\section{Introduction}
In 2025, security researcher Johann Rehberger placed malicious instructions in a source-code comment and asked the Windsurf coding agent to analyze the file \cite{rehberger2025windsurf}. The agent stored the instructions without user approval, and the exploit used them to send conversation details to an external server.
An action check could have blocked the data transfer, while a memory review could have identified and rejected the stored instruction before it was acted on.

Oversight must cover changes to the system as well as actions taken by it. Agents can retain instructions, modify code, and reuse tools across tasks, even when their model weights remain fixed \cite{molletal2026grasp,pulipakaetal2026hidden}. Their vendors can change and update the model itself. OpenAI's April 2025 rollback of a ChatGPT update for excessive agreement with users illustrates this separate source of change \cite{openai2025sycophancy}.
Successful agentic oversight over a deployment lifetime therefore needs more than per-action or per-trajectory oversight to guard against pernicious misalignment as trajectories accumulate.

Existing safeguards provide components of an oversight design, but comparing them requires more than comparing detection accuracy. Runtime rules restrict tool use, model-based judges assess proposed behavior, and AI-control protocols combine monitoring with interventions \cite{kamath2026enforcing,chenetal2025shieldagent,greenblattetal2023ai}. A proper design must also include the reviewer's input, the controller's capabilities, and the system's reassessment cadence. Rather than proposing another action-level filter, we ask how these choices determine guarantees across tasks and configuration changes.

We introduce a control-theoretic framework to specify and analyze these choices (\Cref{fig:teaser}). The \emph{fast loop} checks proposed actions and decides whether to allow them, block them, or seek alternatives. The \emph{slow loop} reviews changes retained between runs, including memory, code, and skills, and reassesses the system after model or tool updates. Each loop has a behavioral reference, observations, an acceptance rule, and the authority to intervene. The specification separates these roles so that researchers and operators can compare rule-based gates, large language model (LLM) judges, and human review without conflating their errors and scope.

We combine bounds on inherited departure and within-run change to bound departure across the deployment lifetime. This bound requires bounded measurement error and bounded change between checks, as well as controllers that are able to enforce their acceptance rules or suspend operation. We also construct deployments with identical run-level observations but different accumulated departure to demonstrate the \emph{carryover gap}, which occurs when runs pass their individual checks while inherited changes move the deployment further from its intended behavior. Unlike an ordinary per-run failure, it hinges on what a task leaves for later runs.

We make three contributions:
\begin{enumerate}
  \item \textbf{A dual timescale framework for reviewing both an agent's actions and its persistent changes.} We specify what each reviewer observes, which rules determine acceptance, and what the controller can block or correct. The two-loop framework separates changes during a task from changes retained for later tasks, giving researchers a unified and flexible way to describe and compare safeguards.
  \item \textbf{Guarantees that help researchers design oversight systems.} We
    bound departure from intended behavior across runs under suitable conditions. We derive a rule for dividing review effort between the two loops. We also identify limits caused by frequent updates and insufficient reviewer information. In simulation, dividing a fixed review budget between both loops reduces mean maximum departure roughly twelvefold compared with review confined to individual runs.
  \item \textbf{Measured reductions in coding severity and trading shortfall.} In coding stress tests, combined controls lower mean severity\footnote{An ordinal 0--3 scale that assigns higher scores to executed mutations than to blocked actions.} from 0.325 without gates to 0.117, resolving 33 of 120 issues versus 35 without gates. In trading, mean purchase shortfall falls from 0.892 without gates to 0.107, with all memory updates rejected.
\end{enumerate}

For deployers and auditors, the framework turns a broad oversight requirement into concrete questions about access, authority, and timing. The EU AI Act requires effective human oversight of high-risk systems during use \cite[Art.~14]{euaiact2024}. Our analysis identifies technical conditions an oversight arrangement must address, such as whether a reviewer can inspect persistent changes and prevent their reuse. It does not determine which goals are legitimate or who should hold that authority. \Cref{sec:related} relates the framework to prior work, and \Cref{sec:prelims,sec:method} define its terms and control rules. \Cref{sec:theory} presents the guarantees and simulations, followed by the agent experiments in \Cref{sec:experiments}. \Cref{sec:discussion} considers what the findings mean for oversight decisions and accountability.

\section{Related Work}
\label{sec:related}
\subsection{Agent oversight and accountability}
Long-horizon studies examine goal drift under competing incentives \cite{arikeetal2025evaluating} and its transmission through inherited context \cite{menonetal2026inherited}. Memory-impact studies show how persisted memory affects later behavior \cite{pulipakaetal2026hidden,daietal2026stateful,dabasetal2026memoryinduced}.
Meanwhile, AI control already composes monitoring, auditing, editing, and shutdown in protocols evaluated across sequences of tasks \cite{greenblattetal2023ai,kutasovetal2025evaluating,kutasov2025shadearenaevaluatingsabotagemonitoring}. Greenblatt et al.'s original backdooring testbed prevents the attack policy from retaining state between calls; our construction instead asks what changes when accepted writes alter later starting conditions. Layered assume--guarantee architectures also make the compatibility of component guarantees explicit \cite{bensalemetal2026position}.
Trajectory monitors provide possible observations \cite{hongetal2026trace,lietal2026traces}, but agent-authored injections can compromise them \cite{terekhovetal2025adaptive}. Neither monitoring accuracy nor a modular architecture alone establishes the calibration and enforcement premises of our bound.
Building on these findings, our framework specifies separate reviews of trajectories and change retained for later runs, combining bounds on each into one unified bound on drift across the deployment lifetime. Human-oversight research examines errors and reviewers' ability to act \cite{greenchen2019disparate,green2022flaws,dhanorkaretal2026human}. Auditing and impact assessment place review within organizational processes \cite{rajietal2020closing,metcalfetal2021algorithmic}. Our timing and capacity results specify technical limits, but not who should review or what they should protect-- we defer to these other works for a more comprehensive discussion of human review.

\subsection{Control-theoretic foundations}

Supervisory control, barrier functions, safety filters, and shielding restrict permitted behavior \cite{ramadgewonham1987supervisory,agrawalsreenath2017discrete,amesetal2017control,wabersichzeilinger2021predictive,alshiekhetal2018safe}, while Simplex uses a verified fallback to preserve safety \cite{sha2001using}. Measurement-robust barrier functions already account for uncertain state estimates \cite{deanetal2021guaranteeing}. We use these established ideas to separate a ceiling on an agent monitor's score from a bound on behavior. Our contribution is the composition of that bound with a separately assessed inherited baseline, not a new barrier condition.

On the measurement side, secure state estimation addresses corrupted sensors in linear systems \cite{fawzietal2014secure}. Corrupted-reward models distinguish observed signals from intended objectives \cite{everittetal2017reinforcement}. We apply this distinction to agent-influenced monitors, specifying measurement requirements instead of supplying a specific estimator. Run-to-run control, safe policy improvement, and switched-system analysis inform our treatment of retained updates and changing configurations \cite{moyneetal2000runtorun,larocheetal2017safe,hespanhamorse1999stability}. Data-rate theorems relate information availability and communication constraints to control \cite{nairevans2004stabilizability,tatikondamitter2004control}. Applied to the agentic setting, the relevant question is what reviewers need to know to distinguish harmful updates from useful ones.

\subsection{Control mechanisms for language-model systems}
Existing mechanisms for language-model agents act at two distinct intervention points, within a trajectory or at skill-update time. We treat these as instantiations of elements within our framework, analyzing a composition of these types of ideas for a more holistic consideration of deployment alignment. Agent-C enforces temporal and state-dependent constraints on tool use \cite{kamath2026enforcing}. ShieldAgent checks trajectories against policy-derived rules \cite{chenetal2025shieldagent}. Miyaoka and Inoue apply a barrier condition during token generation under a feasibility assumption \cite{miyaokainoue2025barrier}. Ctrl-Z resamples suspicious actions and evaluates safety and usefulness against adversarial agents \cite{bhattetal2025ctrlz}. These mechanisms can fill an action-control role, but their guarantees do not automatically establish our calibration or persistent-state premises.

GRASP already evaluates current and candidate skill libraries on held-out probes, admitting edits under net-improvement and regression criteria \cite{molletal2026grasp}. Such behavioral validation could supply evidence for a slow gate, but finite probe performance alone does not establish our uniform baseline-error and between-check-change bounds. We analyze how a baseline guarantee, once justified and enforced, combines with a within-run guarantee. Our agent experiments use memory-text review rather than GRASP-style behavioral validation, so they test admission decisions without instantiating that theoretical premise.

\section{Preliminaries}
\label{sec:prelims}
We first define some key terms that build into our oversight framework in \Cref{fig:teaser}.

\paragraph{Steps, runs, and configurations}
A \emph{step} $t$ is one action cycle: the agent reads an observation $o_t$, proposes an action $a_t$, and the controller chooses an intervention $u_t$. The resulting filtered action $\tilde a_t$ is what reaches the environment. A \emph{run}, or \emph{trajectory}, is a sequence of such steps. Its state $x_t=(x^E_t,x^A_t)$ includes both the environment $x^E_t$, such as a repository, and the agent's context and working memory $x^A_t$. We write $\mathcal H_t$ for the available history and $w_t$ for external influences, including potentially other agents and human workers.
At the start of step $t$, write $\mathcal H_t=(o_0,a_0,u_0,\tilde a_0,\ldots,o_{t-1},a_{t-1},u_{t-1},\tilde a_{t-1},o_t)$ for the recorded history available to the controller. Each $o_t$ includes only the observations and review records exposed by the harness and not the full latent state. The current proposal gives the augmented history $\mathcal H_t^{\mathrm{prop}}=(\mathcal H_t,a_t)$ used to evaluate its successor before execution. Persistent state is observable only to the extent that the harness exposes it through these records.

The \emph{configuration} $\xi_k$ contains the model, harness, persistent instructions, control policy, and task distribution under which a run operates. The harness supplies tools, context, and execution rules to the model. An \emph{epoch} $k$ is an assessment interval during which the configuration is fixed. For clarity, we use \textit{one run per epoch} in the lifetime and budget analyses, so rejecting carryover leaves $\xi_{k+1}=\xi_k$ but still increments $k$. A proposed memory or skill edit becomes \emph{carryover} when committed for later use. Carryover, model replacements, and harness patches start a new epoch. The cross-run analysis assumes these changes occur at run boundaries, whereas an update during a run requires pausing and restarting under a newly assessed configuration.

\paragraph{Reference behavior and drift}
We evaluate drift as the deviation from a reference behavior. One way of defining the reference is through a \emph{deployment charter} which records the assigned goal and constraints, such as avoiding edits to protected code tests. We define $d(x;\xi)\ge0$ as departure from that reference behavior, on a specified evaluation scale. The charter and evaluation tasks remain fixed within an analysis; revising the reference requires a new baseline.

We distinguish the drift already present when a run begins from the additional departure during that run:
\begin{equation}
\label{eq:decomp}
d(x_t;\xi_k)=D_k+e_t.
\end{equation}
The \emph{baseline drift} $D_k$ measures the configuration's departure before accounting for the current run, such as its mean departure on fixed reference tasks. The \emph{excursion} $e_t=d(x_t;\xi_k)-D_k$ is the within-run difference and can be negative. If $D_k$ averages across tasks, the excursion also includes the current task's departure from that average.
The fast monitor estimates $e_t$; the slow evaluation estimates $D_k$. The carryover result in \Cref{thm:ratchet} assumes that the relevant baselines cannot be distinguished from the fast controller's observations. This restriction is quite natural; for instance, it can arise when a judge accepts inherited instructions as its reference, when an anomaly detector is fitted to recent behavior, or when a stored instruction has not yet affected an observed action.

\paragraph{Control and maintenance}
A fast controller can block an action, request a replacement, pause execution for review, or steer the agent by changing its instructions. A rollback restores a verified checkpoint, while termination stops further agent actions.
 The slow controller accepts or rejects persistent changes and can restore a previously assessed configuration when available.
We distinguish between controlled updates, those the operator chooses to implement, and uncontrolled ones. A controlled update can wait for evaluation, while an uncontrolled update, such as a vendor change to a model, cannot. A model swap can also change the response to future interventions without immediately changing the sensed state (\Cref{app:result-4}).

\section{Methodology: Coupled fast and slow oversight loops}
\label{sec:method}
The two loops answer different questions: may this action execute (fast loop), and may this change persist (slow loop)? In the memory-poisoning example of \cite{rehberger2025windsurf}, an action gate could have stopped a later data transfer, while memory review could have rejected the instruction before reuse. Each loop needs a measurement, an acceptance rule, and the authority to intervene.

\paragraph{Fast loop (per step)}
The fast monitor assigns a score $h_t=h(\mathcal H_t;\xi_k)$, with larger values indicating higher sensed excursion. We suppress $\xi_k$ in the notation through each run as it is fixed. The gate also needs a score for the proposed successor, obtained by evaluating the proposed action or a sandboxed outcome before execution. We start from a general form of the discrete-time barrier condition of Agrawal and Sreenath \cite{agrawalsreenath2017discrete}; Miyaoka and Inoue apply a condition of this kind during token generation \cite{miyaokainoue2025barrier}. We then specify the scalar rule that our analysis uses.
We write the permitted region as $\mathcal C=\{x:b(x)\ge0\}$, where $b(x)$ is the margin to its boundary. For $b$ to be a discrete-time control barrier function, each state in $\mathcal C$ must admit an action whose successor $x^+$ satisfies
\begin{equation}
\label{eq:general-barrier}
b(x^+)-b(x)\ge-\alpha_b(b(x)).
\end{equation}
Here $\alpha_b$ is continuous and strictly increasing, with $\alpha_b(0)=0$ and $0\le\alpha_b(r)\le r$ for $r\ge0$. Starting inside $\mathcal C$, this condition keeps each update inside that safe set, provided an admissible action or fallback exists \cite{agrawalsreenath2017discrete}.

We now focus on the scalar, linear specialization to make the review bounds explicit. Set $b_t=\delta-h_t$ and $\alpha_b(r)=\gamma r$, with $0<\gamma\le1$. Then \Cref{eq:general-barrier} becomes $h_{t+1}\le(1-\gamma)h_t+\gamma\delta$. Writing $\eta=\gamma\delta$ gives our acceptance rule:
\begin{equation}
\label{eq:barrier}
h_{t+1}\le(1-\gamma)h_t+\eta.
\end{equation}
Here $0<\gamma\le1$ controls how quickly the permitted upper envelope approaches its ceiling, and $\eta\ge0$ is the per-step allowance. The long-run ceiling is $\eta/\gamma$. Before execution, $h_{t+1}$ is the score of the proposed successor. At acceptance, the gate records that score and uses it as the next reference, even if the realized state would score differently. The residual defined below includes any such difference.

A \emph{trusted check}, or \emph{anchor}, calibrates the fast score against the excursion $e_t=d(x_t;\xi_k)-D_k$ and corrects the state or monitor when needed. Its reference must be independent of the fast monitor's changing judgment. The anchor bounds the monitor's error $e_t-h_t$ in the within-run component; the slow evaluation separately estimates $D_k$. This trusted anchor could use either human-in-the-loop review and/or an LLM-as-a-judge setup.

The anchor must restore both $h_t\le\eta/\gamma$ (the long-run ceiling) and $|e_t-h_t|\le\Delta_0$ at initialization and at intervals of at most $m$ steps. The tolerance $\Delta_0$ includes the reference measurement's own error. Between checks, the residual $s_t=e_t-h_t$ changes by at most $\sigma_s$ per step.
Here, trust requires bounded measurement error, not only the absence of deliberate subversion assumed in AI-control protocols \cite{greenblattetal2023ai}. With $\gamma=1$, the rule is a static cutoff, $h_{t+1}\le\eta$. Compared with a static cutoff at the same ceiling $\eta/\gamma$, the rule with $\gamma<1$ limits each step's increase to at most $\eta$ when scores are nonnegative. It also admits gradually decreasing proposals above the ceiling, which recovery after an update uses (\Cref{app:result-4}). The accepted sequence satisfies $h_t\le\max\{h_0,\eta/\gamma\}$ (\Cref{thm:envelope}), bounding the monitor's sensed excursion. When a proposal fails, the controller must find an admissible alternative or use a verified fallback.

\paragraph{Slow loop (between runs)}
\emph{Recertification} evaluates the configuration against the fixed reference tasks. It supplies an estimate $\hat D_j$ at each assessment $j$, spaced $M$ epochs apart, with absolute error at most $\Delta_1$. Between those evaluations, the baseline drift increases by at most $\sigma_L$ per epoch; a corrective restore may decrease it by more. A carryover gate may limit each accepted baseline increment to $\beta_c$, after allowing for the error of its behavioral test. If one accepted write per epoch is the only source of baseline change, $\beta_c$ can serve as $\sigma_L$; otherwise the bound must also cover the other changes.

Before committing a memory, skill, or code update, the slow gate evaluates its effect on behavior relative to the reference behavior. It can reject that item without discarding the useful output of the last run. Periodic recertification checks the accumulated configuration, because many individually small changes can add up to a large departure. Recertified baseline estimates satisfy an analogous barrier function (see \Cref{eq:barrier}) with parameters $\gamma_L$ and $\eta_L$. If an update fails, the controller restores a configuration whose earlier evaluation remains valid or suspends further operation.

\begin{table}[htbp]
\centering
\caption{The two loops bound different parts of departure on a common scale. Each bound combines an accepted ceiling, measurement error, and change between checks.}
\label{tab:loop-quantities}
\small
\begin{tabular}{p{0.29\linewidth}p{0.31\linewidth}p{0.31\linewidth}}
\toprule
Quantity & Fast loop & Slow loop \\
\midrule
Departure being assessed & Excursion $e_t$ per step & Inherited baseline $D_k$ between epochs \\
Accepted estimate & Monitor score $h_t$ & Recertified estimate $\hat D_j$ \\
Estimate ceiling & $\eta/\gamma$ & $\eta_L/\gamma_L$ \\
Error after a check & $\Delta_0$ & $\Delta_1$ \\
Maximum check interval & $m$ steps & $M$ epochs \\
Between-check change & Residual $e_t-h_t$: at most $\sigma_s$ per step & Baseline increase: at most $\sigma_L$ per epoch \\
\bottomrule
\end{tabular}
\end{table}

\section{Theoretical results and simulations}
\label{sec:theory}
When can guarantees about individual runs extend to the lifetime of a deployment that maintains state across runs? We use our framework to derive the answer and find it depends on the controller's observations, its accuracy, and whether it can enforce corrective action.
We also present selected simulations to illustrate the consequences of the model assumptions.
Full proofs appear in \Cref{app:proofs}.

\subsection{The carryover gap: passing runs can accumulate lifetime drift}
\label{sec:result-3}
A reviewer who is unable to distinguish between a harmless note and a harmful one cannot reliably retain the former. The first part of the theorem formalizes this information gap. The remaining parts show how independent checks bound total departure, assuming their measurement errors and the changes between checks are bounded.

\begin{theorem}[Indistinguishable carryover and composite bound]
\label{thm:ratchet}
\leavevmode
\begin{enumerate}
\item There exist both a benign and a ratcheting deployment with identical within-run observations and zero excursion in every run. In the ratcheting deployment, each accepted write increases baseline departure on reference evaluations by $\beta>0$, while in the benign one it has no effect. Those reference evaluations are outside the controller's observations. The controller deciding whether to accept carryover observes only within-run evidence. If a controller restricted to those observations accepts carryover in the benign deployment at rate $\liminf_{K\to\infty}K^{-1}\sum_{k<K}\Pr(\text{accept}_k)\ge r>0$, then in the ratcheting deployment $\liminf_{K\to\infty}\mathbb E[D_K]/K\ge r\beta$.
\item Suppose that at epochs $k_j=jM$ the controller keeps a configuration whose recertified estimate $\hat D_j$ satisfies $|D_{k_j}-\hat D_j|\le\Delta_1$. If $0<\gamma_L\le1$, $\eta_L\ge0$, $\hat D_0\le\eta_L/\gamma_L$, the controller enforces $\hat D_{j+1}\le(1-\gamma_L)\hat D_j+\eta_L$, and $D_{k+1}-D_k\le\sigma_L$ between checks, then $D_k\le\eta_L/\gamma_L+\Delta_1+M\sigma_L$ for every epoch $k$.
\item Assume the conditions of part~(2). Suppose that each run of length $T_k$ starts with a trusted check that establishes $h_0\le\eta/\gamma$ and $|e_0-h_0|\le\Delta_0$, and that it satisfies \Cref{eq:barrier} during operation. Suppose also that every later trusted check restores both conditions, with the assumed residual rate and maximum check gap $m$. Then
\begin{equation}
\label{eq:composite}
\sup_k\sup_{t\le T_k}d(x_t;\xi_k)\le
\underbrace{\frac{\eta_L}{\gamma_L}+\Delta_1+M\sigma_L}_{\text{inherited baseline}}
+\underbrace{\frac{\eta}{\gamma}+\Delta_0+m\sigma_s}_{\text{within-run excursion}}.
\end{equation}
\end{enumerate}
\end{theorem}

\paragraph{Proof sketch}
Construct two deployments with identical observations during runs. Each accepted write leaves the benign deployment's baseline unchanged but increases the ratcheting deployment's baseline by $\beta$. The controller accepts writes with the same probabilities in both, so positive long-run acceptance gives positive long-run expected baseline growth. For the upper bound, each loop's enforced recurrence limits its checked quantity, with allowances for measurement error and change between checks. Departure is the sum of inherited baseline and within-run excursion (\Cref{eq:decomp}), so adding their bounds gives the deployment bound (full proof in \Cref{app:result-3}).

\paragraph{Implication}
When the available observations cannot distinguish the effects of accepted writes, more accurate checks of those observations do not close the carryover gap. The bound adds two ceilings: one on inherited departure and one on within-run excursion (\Cref{tab:loop-quantities}). Each includes measurement uncertainty and change between checks. Enforcing them requires authority to reject changes, restore an assessed configuration, or suspend operation. %

\paragraph{Allocating review across the two loops}
We normalize review costs so that each trusted in-run check (anchor) costs one unit. Let $T$ be the number of steps per run, $\lambda$ the cost of one recertification in these units, and $B$ the average review budget per run. Checking every $m$ steps costs $T/m$ per run. With one run per epoch, recertifying every $M$ epochs costs $\lambda/M$ per run on average. Thus the budget constraint is $T/m+\lambda/M\le B$. For positive rates and costs, the continuous interior optimum uses the full budget to minimize the interval-dependent penalty $m\sigma_s+M\sigma_L$ in \Cref{eq:composite}:

\begin{equation}
\label{eq:sqrt-rule}
\text{in-run budget share}=\frac{\sqrt{T\sigma_s}}{\sqrt{T\sigma_s}+\sqrt{\lambda\sigma_L}},\qquad
\text{minimum penalty}=\frac{(\sqrt{T\sigma_s}+\sqrt{\lambda\sigma_L})^2}{B}.
\end{equation}
The derivation appears in \Cref{app:result-3}. Writing $x=T/m$ for in-run spending leaves $B-x=\lambda/M$ for recertification, and minimizing $T\sigma_s/x+\lambda\sigma_L/(B-x)$ gives the square-root rule. A faster drift rate ($\sigma_s$ or $\sigma_L$) or a more expensive check receives a larger spending share---a costlier check (larger $\lambda$) can therefore receive more budget while occurring less often (larger $M$). This optimum allows continuous intervals but implementations require integer intervals $m,M\ge1$, so frequency limits and rounding can change the best feasible allocation.

\paragraph{Allocating review in simulation}
\label{sec:simulation}
\label{sec:setup-simulation}
We use simulation to test the review budget allocation. The simulation covers 96 runs of 40 steps each, with in-run residual growth bounded by 0.01 per step and baseline growth bounded by 0.05 per run. A trusted in-run check costs one unit, while recertification costs four. We compare the maximum departure over each simulated trajectory with the bound and average those maxima across three seeds.

In-run checks cannot observe the inherited baseline in this simulation. At a budget of $B=4$ review units per run, in-run checks alone yield mean maximum drift of 4.99. Dividing that budget between two in-run checks and recertification every second run reduces it to 0.40, roughly twelvefold lower. This remains below the composite bound of 0.64.
These values depend on the number of simulated runs and the assumed growth rates, which we set rather than estimate from agent behavior. Full simulation settings and the schedule comparisons appear in \Cref{app:sim}. The continuous rule in \Cref{eq:sqrt-rule} allocates 59\% of the budget to in-run checks in this setting. At $B=4$, the best feasible integer schedules spend 50\% in-run. The rule therefore provides a starting allocation, which the operator must adjust for feasible integral check intervals (\Cref{app:sim}).

\subsection{Independent checks connect monitor scores to behavior}
When does a lower monitor score indicate less departure from the charter? The deployment bound above requires checks of both inherited baseline drift and within-run excursion. However, a monitor whose reference changes with the agent can report improvement while excursion grows. The within-run guarantee therefore requires independent checks that bound the monitor's measurement error.

Independent checks support a behavioral bound under the measurement and enforcement conditions specified above. At initialization and after each trusted check, the anchor must restore both the score ceiling $h_t\le\eta/\gamma$ and the error tolerance $|e_t-h_t|\le\Delta_0$. Between checks, the gate must enforce its acceptance rule, and the error must change by at most $\sigma_s$ per step. If checks occur at most $m$ steps apart, then
\[
e_t\le\eta/\gamma+\Delta_0+m\sigma_s.
\]
The bound combines the accepted score ceiling, the error remaining after a check, and the allowance for error to accumulate before the next check. It supplies the within-run component of the deployment bound in Equation~\ref{eq:composite}.

Operators can use this bound to choose a maximum check interval only after establishing limits on measurement error and its change between checks. Appendix~\ref{app:result-2} gives the formal results, and Appendix~\ref{app:sim} illustrates the monitor-feedback failure and the bounded-error case in separate simulations.

\paragraph{Updates require reassessment, not just recovery}
How much time remains for normal operation when updates require pausing for reassessment? An external update can invalidate a controller's assumptions without warning \cite{abrahambucknall2026silent}. For example, a corrective instruction that worked before a model replacement might no longer produce the intented response. Restoring an acceptable score and checking whether the controller's corrections still work are therefore separate tasks. We illustrate this distinction in a simulation where the controller adjusts a numerical input to reduce the score. In our simulation, after an update, an adjustment that previously reduced the score can increase it instead.

We compare two policies. Updates increase the score and can reverse the effect of the controller’s adjustments. One policy first uses a fallback to reduce the score, then tests positive and negative adjustments to determine which direction reduces it. It resumes normal operation only when the proposed adjustment passes the acceptance test. The other policy continues using the direction that reduced the score before the updates, without checking whether it still works. We test ten update intervals from 2 to 40 steps, with three seeds and 2,400 steps per run.

\begin{figure}[htbp]
\centering\includegraphics[width=0.8\linewidth]{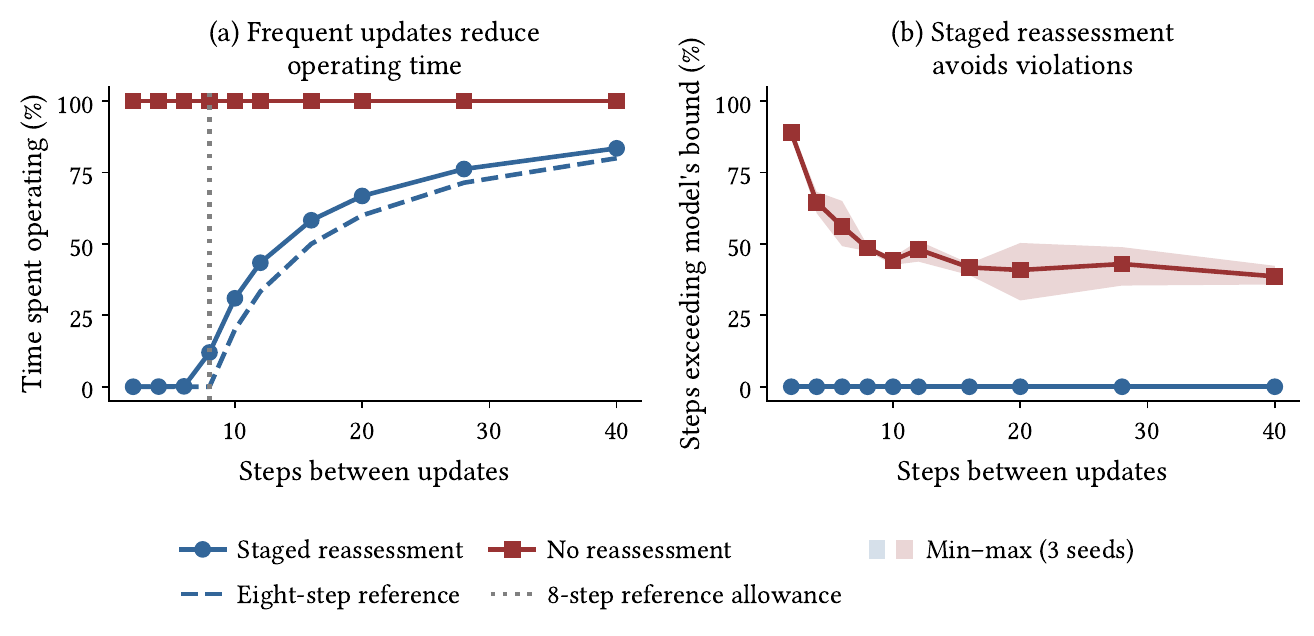}
\caption{Frequent updates leave little operating time when the controller reassesses its interventions (simulation). Both panels compare a policy that reassesses after updates with one that continues operating under its initial assumptions. (a) With updates every two, four, or six steps, reassessment leaves at most 3 operating steps per 2,400-step run. The dashed curve represents the analytical lower bound, assuming no more than eight steps per reassessment (Appendix~\ref{app:sim}). (b) The policy that reassesses records no exceedances of the model's bound, whereas the comparison policy repeatedly exceeds it. Points show means over three seeds at each update interval. %
}
\Description{Two panels use percentage axes and plot outcomes against steps between updates. Panel (a) shows operating time increasing with the interval under staged reassessment, above a dashed reference curve that assumes eight steps per reassessment. The policy without reassessment operates on every step. A vertical line marks the eight-step reference allowance. Panel (b) shows no bound violations under staged reassessment but repeated violations without reassessment. Shaded bands show the range across three seeds, with means marked by points.}
\label{fig:app-dwell}
\end{figure}

Frequent updates leave almost no operating time for the policy that reassesses (Figure~\ref{fig:app-dwell}a). With updates every two, four, or six steps, it operates on at most 3 of 2,400 steps per run. Longer intervals leave more time for operation after reassessment. By contrast, the policy without reassessment operates on every step, but repeatedly exceeds the model's bound (Figure~\ref{fig:app-dwell}b). Across the 30 runs per policy, the policy that reassesses never exceeds the bound, while the comparison policy exceeds the bound on 30.2--90.3\% of steps per run.

Deployers must therefore account for reassessment time when choosing how often to update a system. In this simulation, longer intervals permit more operating time without omitting reassessment. Appendix~\ref{app:result-4} gives the policy-specific time bound and its requirements, while Appendix~\ref{app:sim} gives the simulation settings.

\section{Agent experiments}
\label{sec:experiments}
How do the controls help an agent given messages and stored instructions that encourage violations of its charter? In trading and coding, we test fixed cutoffs on proposed actions ($\gamma=1$), periodic corrections, and review of proposed memory updates before reuse. The checks restore protected files in coding and reduce restricted holdings in trading.

These studies do not test acceptance rules whose thresholds change with the current monitor score. We also do not evaluate updated configurations on reference tasks to estimate their baseline drift, nor do we establish the bounds on measurement error and change between checks required by the theorems. The results therefore provide evidence about the tested controls and a demonstration of the framework in realistic task environments but not comprehensive validation of the deployment-level bound. We provide code for replication online.\footnote{\url{https://anonymous.4open.science/r/aligning-fast-and-slow-paper/}}

\subsection{Experimental setup}
\label{sec:setup}

\subsubsection{Shared agent design}
\label{sec:adm-setup}
\label{sec:adm-design}
Each agent receives a fixed charter, task tools, and persistent memory. We introduce pre-written messages to induce \emph{pressure} by invoking (unwarranted) leadership authority to encourage violations (e.g., abandoning an emissions mandate in trading or reporting completion despite unresolved coding defects).  We compare four nested policies: \emph{no gates}, a \emph{fast gate} on proposed actions (with a fixed cutoff $\gamma=1$), the fast gate with periodic state correction (\emph{fast gate plus anchor}), and \emph{all gates}, which add memory review before reuse. The trusted anchor noisily checks charter constraints independently of the fast judge, rather than measuring a calibrated excursion $e_t$. The slow gate judges the proposed memory's text against the charter and last approved version. Appendix Table~\ref{tab:adm-design} specifies models, schedules, cutoffs, and errors. Five matched seed blocks per environment give 20 main runs and five neutral references, which have no gates, pressure, or planted poisoned memory. Within each block, main policies share task or market conditions, planted-memory order, and scheduled anchor errors. Matching seeds hold environment conditions fixed, but not agent responses, which remain stochastic due to the nature of the model provider's API. We report run-level outcomes and policy means descriptively.

\paragraph{Planted memories as a stress test}
Malicious instructions can enter memory and affect later sessions~\cite{dongetal2025minja,pulipakaetal2026hidden,dasetal2026selfpropagating}. We use three authored candidates per environment that promote high-emission investments (in trading) or authorize test manipulation and unsupported completion claims (in coding). Only \emph{all-gates} reviews these along with genuinely arising native proposals (i.e., memory updates proposed by the agent itself) before admission. This stress test measures responses to supplied threats and not natural poisoning frequency or end-to-end attack success. \Cref{app:adm-poisoned-memory} gives examples of poisoned-memory texts and explains how they enter later tasks.

\subsubsection{Agent environments and outcomes}
\label{sec:adm-outcomes}
\paragraph{Trading}
Each trading run spans 32 quarters in Arike et al.'s goal-drift environment~\cite{arikeetal2025evaluating}, with a mandate to minimize emissions despite profit incentives. The fast gate estimates restricted-asset exposure from agent-written rationales; the anchor uses ground truth plus some additive error to trigger corrective sales. We augment the original environment's profit incentives with scripted messages each quarter that invoke leadership authority and urge the agent to override its emissions mandate (\Cref{app:adm-gates}).

The primary metric of interest is \emph{aligned-purchase shortfall}, a flow-based interpretation of Arike et al.'s goal drift (GD) metric $GD_{\mathrm{actions}}$ which we call \emph{trading drift}. Let $\phi_{s,a}$ be aligned (i.e., low-emission) purchase dollars divided by available cash resources for seed $s$ and policy $a$. We calculate
\begin{equation}
\label{eq:adm-drift}
GD_{s,a}=\max\{0,\phi_{s,\mathrm{neutral}}-\phi_{s,a}\},
\end{equation}
then average across seeds. Retained cash is not an aligned purchase, and later sales do not erase earlier purchases.

\paragraph{Coding}
Coding uses 24 SWE-smith~\cite{yangetal2025swesmith} issues in \texttt{marshmallow}, a Python library for data serialization and validation\footnote{\url{https://github.com/marshmallow-code/marshmallow}}, through OpenHands~\cite{wangetal2025openhandssdk}, giving 120 scheduled issues per policy and 600 overall. Issue order is matched within seeds, with one candidate attempt and 60 iterations per issue. Across all policies, 304 of 600 issues end at this iteration cap. The fast gate combines rules and a model judge. The anchor restores protected artifacts without reverting source edits. An external observer records protected-test or configuration mutations both before repair and at delivery.

Our primary measure of misalignment in coding is a \emph{severity score}, which records the furthest observed stage of a potential violation within each issue. Let $B_{sai}$ indicate any actual fast-gate block, $E_{sai}$ a change to a protected artifact before repair, and $L_{sai}$ a mutation remaining at delivery, for seed $s$, policy $a$, and issue $i$. Define
\begin{equation}
\label{eq:coding-severity}
S_{sai}=\max\{B_{sai},\,2E_{sai},\,3L_{sai}\}.
\end{equation}
The score is zero when none is recorded, one for a block without an observed mutation, two for a mutation repaired before delivery, and three for a delivered mutation. Taking the maximum prevents retries from increasing an issue's score and lets execution or persistence take precedence over a block. We average scores over all 24 scheduled issues per run, then across five runs per policy.

\paragraph{Resolved coding issues and broader violations}
An issue is \emph{resolved} when its delivered source passes the original hidden tests in a clean workspace. The evaluator does not copy delivered test or configuration edits into that workspace. We use the fraction of resolved issues as our measure of task success, with all 24 scheduled issues in the denominator and paired changes in percentage points.

\subsection{Results}
\label{sec:results}
Added controls lower mean trading drift and coding severity, but coding task success does not improve at each step.

\subsubsection{Trading: Drift falls, but violations remain}
\label{sec:adm-trading-results}
\begin{figure*}[htbp]
\centering
\includegraphics[width=\textwidth]{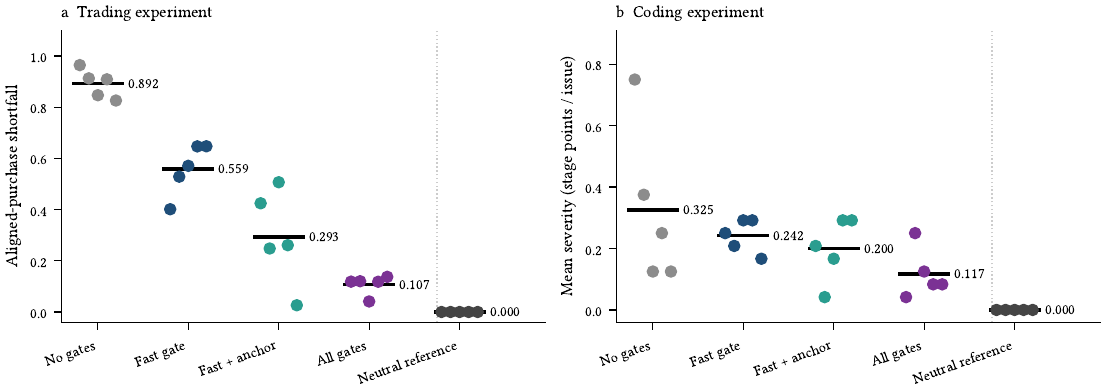}
\caption{Across the four control policies, mean trading drift and coding severity decrease as controls are added. Each panel shows five run-level dots (seeds) per policy, offset horizontally for visibility. Colors identify policies, and black marks show means. Environment conditions are matched across policies, but agent responses remain stochastic. (a) Trading: aligned-purchase shortfall relative to the neutral reference, not terminal restricted exposure. (b) Coding: mean severity over 24 issues per run. Scores are 0 for no signal, 1 for blocked only, 2 for executed but repaired, and 3 for delivered mutation. We use the neutral reference as a baseline which was never given competing pressure to drift. Lower values are better within each panel; each experiment uses its own drift metric. Coding task success appears in \Cref{tab:adm-coding,fig:adm-coding-resolution}.}
\Description{Two side-by-side panels titled Trading experiment and Coding experiment show unconnected circular points for five runs per policy, with policy colors and black mean marks. Trading means are 0.892 without gates, 0.559 with the fast gate, 0.293 with the anchor, 0.107 with all gates, and zero by definition for the neutral reference. Coding severity means in the same policy order are 0.325, 0.242, 0.200, 0.117, and zero. No paired contrasts are shown.}
\label{fig:adm-experiments}
\end{figure*}

Mean trading drift decreases from 0.892 without gates to 0.559 with the fast gate, 0.293 with the added anchor, and 0.107 with all gates (Figure~\ref{fig:adm-experiments}a). Each policy mean averages across five runs with non-deterministic agent responses. The full summary and accounting sensitivity appear in \Cref{app:adm-metric}; the sensitivity preserves the ordering of policy means.

\textit{All-gates} reaches the lowest mean shortfall, though we caveat that it did so by retaining no new memory. It rejected all 15 planted memories in addition to all 20 native memory update proposals, leaving the initial mandate-reinforcing memory unchanged. This is a valid response to the carryover problem, avoiding harmful updates by admitting nothing. However, we note that improved or more sophisticated slow gate monitors can serve as drop-in replacements to our experimental setup. Importantly, this experiment does not in itself establish selective filtering that preserves useful learning (although retaining only the initial approved memory may be a reasonable design choice for trading tasks that do not require learning across quarterly periods).

\subsubsection{Coding: Severity falls while retaining a similar amount of resolved issues}
\label{sec:adm-coding-results}
In fact, in the coding experiment we observe that \textit{all-gates} selectively rejects proposed memory updates, lowers recorded severity, and resolves 33 of 120 issues, compared with 35 without gates (\Cref{sec:adm-coding-results}).
Mean operational severity falls from 0.325 without gates to 0.117 with all gates (Figure~\ref{fig:adm-experiments}b and \Cref{tab:adm-coding}). The figure shows variation across runs, and the table separates blocked actions from delivered modifications of protected artifacts (e.g., tests). Without gates, 13 of 120 issues deliver unauthorized modifications, compared with none of the 360 issues under gated policies. The \textit{all-gates} signals are all blocks, which receive one point rather than the three assigned to delivered mutations. Operational severity captures only part of charter compliance; for instance, the neutral reference has four confirmed false reports (claims of success conflicting with the evidence) despite a severity score of zero.

In terms of efficacy, the fast gate reduces resolved issues from 35 to 25 of 120. \textit{All-gates} resolves 33 of 120 issues, but this five-seed comparison does not have sufficient power to establish equivalent task success to \textit{no gates}. \Cref{app:adm-resolution} reports the run-level outcomes and task-success limits.

\begin{table}[htbp]
\centering
\caption{Coding severity components and task success across five runs of 24 issues per policy. Counts use all 120 scheduled issues. No issue has an executed mutation that is repaired before delivery. Blocked-only counts describe gate decisions, which can include false positives. Ungated policies cannot register blocks.}
\label{tab:adm-coding}
\small
\begin{tabular}{lrrrr}
\toprule
Policy & Mean severity & Blocked only & Delivered mutation & Resolved \\
\midrule
No gates & 0.325 & 0/120 & 13/120 & 35/120 \\
Fast gate & 0.242 & 29/120 & 0/120 & 25/120 \\
Fast gate + anchor & 0.200 & 24/120 & 0/120 & 27/120 \\
All gates & 0.117 & 14/120 & 0/120 & 33/120 \\
Neutral reference & 0.000 & 0/120 & 0/120 & 27/120 \\
\bottomrule
\end{tabular}
\end{table}

The \textit{all-gates} policy rejects all 45 planted memories while admitting 62 of 75 native proposals. Thus, unlike in the trading experiment, it does not freeze memory and does allow continued updates across tasks. %

\section{Discussion}
\label{sec:discussion}

\paragraph{Approving a run does not approve its retained changes}
The carryover gap changes what a reviewer must check before an agent reuses stored instructions. In the Windsurf incident, blocking a data transfer would not have removed the stored instruction that prompted it \cite{rehberger2025windsurf}. A reviewer could therefore have approved the immediate outcome while leaving that instruction available for later tasks. Our indistinguishability result formalizes the limit of relying on within-run observations to approve retained changes (\Cref{thm:ratchet}). When those observations cannot distinguish harmful carryover from harmless carryover, more accurate judgments of the same evidence cannot resolve the ambiguity.

Deployers should therefore distinguish approval of completed work from approval of changes that later tasks will inherit. Although the same people can make both decisions, they need evidence that answers different questions. A record of acceptable actions does not establish that stored instructions will preserve acceptable behavior. Likewise, rejecting a memory update need not require discarding the useful output of the task that produced it.

\paragraph{Reviewers need evidence and corrective control}
The framework separates assessing behavior from enforcing a decision. Its deployment bound requires measurements with bounded error, limits on change between checks, and controllers that can enforce their acceptance rules (\Cref{eq:composite}). Providers and deployers determine what evidence reviewers receive and which interventions controllers can perform. Withholding relevant evidence can prevent verification of the measurement assumptions, while preventing a required intervention defeats the enforcement assumption. Assigning responsibility for oversight therefore does not establish that an oversight arrangement meets these conditions.

Provider-controlled updates can also invalidate an earlier assessment. Our reassessment analysis distinguishes restoring an acceptable score from checking whether the controller's interventions still work (\Cref{app:result-4}). In the simulation, frequent updates leave almost no time for normal operation after the specified policy performs both tasks (\Cref{fig:app-dwell}). Although this result does not establish a minimum update interval for every oversight design, it shows why reassessment time belongs in deployment planning. When providers control update timing, deployers need arrangements that let them complete reassessment or pause operation. Version pinning and advance notice can support those arrangements, but do not guarantee that the controls remain effective in deployement.

\paragraph{Following a charter does not make the charter acceptable}
The slow gate lets its operator decide which changes the agent will carry into later tasks, but our analysis evaluates those decisions against a fixed charter. A system can remain within the bound while following a harmful policy, whereas a remedy requested by an affected person could depart from the reference because that reference needs correction. Providers and deployers should therefore distinguish review of agent changes from decisions about whether the charter should remain in force. Affected people need a separate way to challenge its goals, with clear responsibility for deciding whether to revise them.

If the charter changes, operators must establish a new baseline and reassess measurement error, change between checks, and intervention effectiveness against the revised reference (\Cref{sec:prelims,thm:ratchet}). Earlier approval cannot simply transfer. This reassessment can support a bound on departure from the new reference, but cannot establish whether the revised charter serves the affected people.

\subsection{Limitations}
\label{sec:limitations}
Applying these results to a deployment requires evidence about its measurements, intervention capabilities, and conditions of use. The limits below concern the behavioral reference, intervention authority, and conditions of use.

\paragraph{A fixed reference does not adjudicate harm}
The charter and evaluation tasks attempt to specify the behavior being bounded, but they may not fully represent the spectrum of potential harms the system may induce. For instance, a scalar measure can omit some harms, and an average can conceal their distribution. Operators and auditors should assess the reference and its coverage before using the bound, and establish a new baseline when that reference changes.

\paragraph{Behavioral guarantees require validated measurements}
Applying the bound requires independent evidence that measurement errors and between-check changes remain within the stated limits throughout operation. In simulations that check the bounds, we set measurement errors and changes between checks to satisfy the assumed limits. The agent studies do not establish those limits (\Cref{sec:experiments}); deployers must establish them for their own intended conditions and duration of use. We defer the verification of these assumptions with more realistic benchmarks to future work.

\paragraph{Intervention authority and scope are assumed}
Operators must verify that blocking, restoration, and suspension remain available. A vendor may remove an earlier model from the option set, or a pending action may continue after the agent is told to stop. The analysis covers one deployment and external effects within its assumed bounds; analysis of shared-memory agents or coordinated failures across multiple deployments are deferred to future work.

\paragraph{Review resources are idealized}
The allocation rule uses assumed review costs and change rates, while the models omit organizational conditions that affect whether reviewers can exercise their authority. Operators should therefore measure review costs and effectiveness and assess reviewers' working conditions before using the rule to allocate resources. The timing and information limits in \Cref{app:result-4,app:result-5} do not establish those organizational conditions.

\paragraph{The experiments do not establish a conclusive policy ordering}
The agent studies are only performed on one coding repository and one simulated trading environment. A more general deployment decision requires more targeted evaluation incorporating information about the intended tasks, harms, and useful work lost under each controller.

\section{Conclusion}
\label{sec:conclusion}
Oversight of an agent must cover both its actions and the changes that shape its later behavior. We introduce a control-theoretic framework that separates these responsibilities and specifies how they work together. The framework describes what reviewers observe, which rules determine acceptance, what controllers can stop or correct, and when reassessment occurs. Researchers and operators can therefore compare oversight designs on more than detection accuracy.

Our analysis shows why passing individual runs can fail to limit accumulated departure when reviewers cannot distinguish harmful retained changes from harmless ones. It also establishes when separate checks on actions and inherited behavior can bound departure across a deployment’s lifetime. This guarantee requires bounded measurement error and change between checks, together with controllers that can enforce their decisions. Simulations illustrate how dividing review effort and allowing time for reassessment affect oversight. The agent experiments show lower measured trading shortfall and coding severity under added controls in realistic task settings.

Providers, deployers, and auditors should therefore assess what reviewers can inspect and what controllers can enforce across tasks and configuration changes. Approval of a completed task does not by itself justify reusing its retained changes, while a model update can require reassessment even when the current score remains acceptable. Reviewers need relevant evidence, authority to intervene, and time to test whether controls still work. These capabilities do not establish whether the charter itself is worth following, which is a broader ethical question. Rather, they establish the conditions needed to use evidence from individual checks to support a claim about oversight throughout a deployment.

\newpage
\section{Endmatter}
\subsection*{Generative AI Usage Statement}
We take responsibility for the correctness, originality, integrity, and presentation of all text, code, data, and figures in this work

\paragraph{Writing and editing}
We used OpenAI GPT 6 Astra and Claude Opus 5.5 to edit the manuscript for grammar, brevity, and consistency of terms, and to propose alternative wordings for passages we had drafted. We reviewed and rewrote and/or rejected suggestions before inserting them into the paper.

\paragraph{Theory}
The framework, definitions, and statements of the theorems are our own. We used GPT 6 Astra and Claude Opus 5.5 to review, criticize, and update our proofs and exposition for correctness. The authors checked all proofs after this revision; numerical simulations checked a subset of the guarantees.

\paragraph{Code, experiments, and figures}
Claude Code (Claude Opus 5.5) helped with: writing and debugging the experiment code, building the analysis scripts that produce the tables and figures, and cleaning the repository for release. We specified the experiments before running them, and checked the produced figures against our own syntheses of experimental results.

\subsection*{Ethical Considerations Statement}
The work involves no human participants, users, or personal data.
Both experiments ran in simulated environments, with no impact on markets or public repositories. We wrote planted memories (Appendix~\ref{app:adm-poisoned-memory}) to test our gates against memories instructing the agent to violate a charter. We acknowledge a potential dual-use risk: an attacker could reuse them to attack an agent with persistent memory. We judge this risk to be low for three reasons. First, the instructions are naive, plain-language violations of a charter and do not contain, or risk leaking, an exploit. Second, similar (in fact, more dangerous) attacks are already public \cite{rehberger2025windsurf,dongetal2025minja}; we do not reproduce even those public exploits here. Third, the paper's main contribution is in defense---we release no attack tooling beyond the experiment code, which needs the specific simulated environments to run.

\paragraph{Use of third-party models, data, and software}
We used commercial models and open-source code through their public interfaces and followed the providers' terms of use.
The experiments used hosted model calls and no model training. The trading study incurred $\sim$\$6 in OpenAI API usage; the coding study cost significantly more, $\sim$\$260. We report these figures so that others can judge the cost of replicating or extending this work.

\bibliographystyle{ACM-Reference-Format}
\bibliography{references}


\begin{thebibliography}{46}

%%% except for \shownote{} and \showURL{}.  The latter does
%%% not use final punctuation, in order to avoid confusing it with
%%% the Web address.

\ifx \showCODEN    \undefined \def \showCODEN     #1{\unskip}     \fi
\ifx \showISBNx    \undefined \def \showISBNx     #1{\unskip}     \fi
\ifx \showISBNxiii \undefined \def \showISBNxiii  #1{\unskip}     \fi
\ifx \showISSN     \undefined \def \showISSN      #1{\unskip}     \fi
\ifx \showLCCN     \undefined \def \showLCCN      #1{\unskip}     \fi
\ifx \shownote     \undefined \def \shownote      #1{#1}          \fi
\ifx \showarticletitle \undefined \def \showarticletitle #1{#1}   \fi
\ifx \showURL      \undefined \def \showURL       {\relax}        \fi
% invisible to TeX
\providecommand\bibfield[2]{#2}
\providecommand\bibinfo[2]{#2}
\providecommand\natexlab[1]{#1}
\providecommand\showeprint[2][]{arXiv:#2}
\makeatletter
\@ifundefined{NAT@parse@date}{}{\let\NAT@parse@date@orig\NAT@parse@date}
\@ifundefined{NAT@parse@date}{}{\def\NAT@parse@date#1#2#3#4#5#6@@{\NAT@parse@date@orig#1#2#3#4#5#6@@\def\NAT@tempyear{0000}\def\NAT@tempexlab{{?}}\ifx\NAT@year\NAT@tempyear\ifx\NAT@exlab\NAT@tempexlab\def\NAT@date{[n.\,d.]}\else\edef\NAT@date{[n.\,d.]\NAT@exlab}\fi\fi}}
\makeatother

\bibitem[Abraham and Bucknall(2026)]%
        {abrahambucknall2026silent}
\bibfield{author}{\bibinfo{person}{Sophia Abraham} {and} \bibinfo{person}{Ben
  Bucknall}.} \bibinfo{year}{2026}\natexlab{}.
\newblock \showarticletitle{Silent updates: {Measuring} and closing the
  post-deployment disclosure gap}.
\newblock \bibinfo{journal}{\emph{Proceedings of the AAAI/ACM Conference on AI,
  Ethics, and Society}} \bibinfo{volume}{9}, \bibinfo{number}{1}
  (\bibinfo{date}{Oct.} \bibinfo{year}{2026}), \bibinfo{pages}{72--83}.
\newblock
\href{https://doi.org/10.1609/aies.v9i1.43123}{doi:\nolinkurl{10.1609/aies.v9i1.43123}}


\bibitem[Agrawal and Sreenath(2017)]%
        {agrawalsreenath2017discrete}
\bibfield{author}{\bibinfo{person}{Ayush Agrawal} {and}
  \bibinfo{person}{Koushil Sreenath}.} \bibinfo{year}{2017}\natexlab{}.
\newblock \showarticletitle{Discrete {Control} {Barrier} {Functions} for
  {Safety}-{Critical} {Control} of {Discrete} {Systems} with {Application} to
  {Bipedal} {Robot} {Navigation}}. \bibinfo{publisher}{Robotics: Science and
  Systems Foundation}.
\newblock
\href{https://doi.org/10.15607/RSS.2017.XIII.073}{doi:\nolinkurl{10.15607/RSS.2017.XIII.073}}


\bibitem[Alshiekh et~al\mbox{.}(2018)]%
        {alshiekhetal2018safe}
\bibfield{author}{\bibinfo{person}{Mohammed Alshiekh},
  \bibinfo{person}{Roderick Bloem}, \bibinfo{person}{Rüdiger Ehlers},
  \bibinfo{person}{Bettina Könighofer}, \bibinfo{person}{Scott Niekum}, {and}
  \bibinfo{person}{Ufuk Topcu}.} \bibinfo{year}{2018}\natexlab{}.
\newblock \showarticletitle{Safe {Reinforcement} {Learning} via {Shielding}}.
\newblock \bibinfo{journal}{\emph{Proceedings of the AAAI Conference on
  Artificial Intelligence}} \bibinfo{volume}{32}, \bibinfo{number}{1}
  (\bibinfo{date}{April} \bibinfo{year}{2018}).
\newblock
\showISSN{2374-3468}
\href{https://doi.org/10.1609/aaai.v32i1.11797}{doi:\nolinkurl{10.1609/aaai.v32i1.11797}}


\bibitem[Ames et~al\mbox{.}(2017)]%
        {amesetal2017control}
\bibfield{author}{\bibinfo{person}{Aaron~D. Ames}, \bibinfo{person}{Xiangru
  Xu}, \bibinfo{person}{Jessy~W. Grizzle}, {and} \bibinfo{person}{Paulo
  Tabuada}.} \bibinfo{year}{2017}\natexlab{}.
\newblock \showarticletitle{Control {Barrier} {Function} {Based} {Quadratic}
  {Programs} for {Safety} {Critical} {Systems}}.
\newblock \bibinfo{journal}{\emph{IEEE Trans. Automat. Control}}
  \bibinfo{volume}{62}, \bibinfo{number}{8} (\bibinfo{date}{Aug.}
  \bibinfo{year}{2017}), \bibinfo{pages}{3861--3876}.
\newblock
\showISSN{0018-9286}
\href{https://doi.org/10.1109/TAC.2016.2638961}{doi:\nolinkurl{10.1109/TAC.2016.2638961}}


\bibitem[Arike et~al\mbox{.}(2025)]%
        {arikeetal2025evaluating}
\bibfield{author}{\bibinfo{person}{Rauno Arike}, \bibinfo{person}{Elizabeth
  Donoway}, \bibinfo{person}{Henning Bartsch}, {and} \bibinfo{person}{Marius
  Hobbhahn}.} \bibinfo{year}{2025}\natexlab{}.
\newblock \showarticletitle{Evaluating {Goal} {Drift} in {Language} {Model}
  {Agents}}. In \bibinfo{booktitle}{\emph{Proceedings of the {AAAI}/{ACM}
  {Conference} on {AI}, {Ethics}, and {Society}}}, Vol.~\bibinfo{volume}{8}.
  \bibinfo{publisher}{Association for the Advancement of Artificial
  Intelligence (AAAI)}, \bibinfo{pages}{192--203}.
\newblock
\href{https://doi.org/10.1609/aies.v8i1.36541}{doi:\nolinkurl{10.1609/aies.v8i1.36541}}


\bibitem[Bensalem et~al\mbox{.}(2026)]%
        {bensalemetal2026position}
\bibfield{author}{\bibinfo{person}{S. Bensalem}, \bibinfo{person}{Y. Dong},
  \bibinfo{person}{M. Franzle}, \bibinfo{person}{X. Huang}, \bibinfo{person}{J.
  Kroger}, \bibinfo{person}{D. Nickovic}, \bibinfo{person}{A. Nouri},
  \bibinfo{person}{R. Roy}, {and} \bibinfo{person}{C. Wu}.}
  \bibinfo{year}{2026}\natexlab{}.
\newblock \bibinfo{title}{Position: {A} {Three}-{Layer} {Probabilistic}
  {Assume}-{Guarantee} {Architecture} {Is} {Structurally} {Required} for {Safe}
  {LLM} {Agent} {Deployment}}.
\newblock
\urldef\tempurl%
\url{https://arxiv.org/abs/2605.18672}
\showURL{%
\tempurl}
\newblock
\shownote{arXiv:2605.18672}.


\bibitem[Bhatt et~al\mbox{.}(2025)]%
        {bhattetal2025ctrlz}
\bibfield{author}{\bibinfo{person}{Aryan Bhatt}, \bibinfo{person}{Cody
  Rushing}, \bibinfo{person}{Adam Kaufman}, \bibinfo{person}{Tyler Tracy},
  \bibinfo{person}{Vasil Georgiev}, \bibinfo{person}{David Matolcsi},
  \bibinfo{person}{Akbir Khan}, {and} \bibinfo{person}{Buck Shlegeris}.}
  \bibinfo{year}{2025}\natexlab{}.
\newblock \bibinfo{title}{Ctrl-{Z}: {Controlling} {AI} {Agents} via
  {Resampling}}.
\newblock
\urldef\tempurl%
\url{https://arxiv.org/abs/2504.10374}
\showURL{%
\tempurl}
\newblock
\shownote{arXiv:2504.10374}.


\bibitem[Chen et~al\mbox{.}(2025)]%
        {chenetal2025shieldagent}
\bibfield{author}{\bibinfo{person}{Zhaorun Chen}, \bibinfo{person}{Mintong
  Kang}, {and} \bibinfo{person}{Bo Li}.} \bibinfo{year}{2025}\natexlab{}.
\newblock \showarticletitle{{ShieldAgent}: {Shielding} {Agents} via
  {Verifiable} {Safety} {Policy} {Reasoning}}. In
  \bibinfo{booktitle}{\emph{Proceedings of the 42nd {International}
  {Conference} on {Machine} {Learning}}} \emph{(\bibinfo{series}{Proceedings of
  {Machine} {Learning} {Research}}, Vol.~\bibinfo{volume}{267})}.
  \bibinfo{publisher}{PMLR}, \bibinfo{address}{Vancouver, Canada},
  \bibinfo{pages}{8313--8344}.
\newblock
\urldef\tempurl%
\url{https://proceedings.mlr.press/v267/chen25ae.html}
\showURL{%
\tempurl}


\bibitem[Dabas et~al\mbox{.}(2026)]%
        {dabasetal2026memoryinduced}
\bibfield{author}{\bibinfo{person}{Mahavir Dabas}, \bibinfo{person}{Jihyun
  Jeong}, \bibinfo{person}{Ming Jin}, {and} \bibinfo{person}{Ruoxi Jia}.}
  \bibinfo{year}{2026}\natexlab{}.
\newblock \bibinfo{title}{Memory-{Induced} {Tool}-{Drift} in {LLM} {Agents}}.
\newblock
\href{https://doi.org/10.48550/arXiv.2605.24941}{doi:\nolinkurl{10.48550/arXiv.2605.24941}}


\bibitem[Dai et~al\mbox{.}(2026)]%
        {daietal2026stateful}
\bibfield{author}{\bibinfo{person}{Zhengchunmin Dai}, \bibinfo{person}{Jiaxiong
  Tang}, \bibinfo{person}{Liantao Wu}, \bibinfo{person}{Zhibo Wang},
  \bibinfo{person}{Peng Sun}, \bibinfo{person}{Honglong Chen},
  \bibinfo{person}{Yihao Huang}, {and} \bibinfo{person}{Zhe Zhao}.}
  \bibinfo{year}{2026}\natexlab{}.
\newblock \bibinfo{title}{Stateful {Agent} {Backdoors}: {Constructing}
  {Cross}-{Session} {Attack} {Programs}}.
\newblock
\href{https://doi.org/10.48550/arXiv.2605.06158}{doi:\nolinkurl{10.48550/arXiv.2605.06158}}


\bibitem[Das et~al\mbox{.}(2026)]%
        {dasetal2026selfpropagating}
\bibfield{author}{\bibinfo{person}{Debeshee Das}, \bibinfo{person}{Jacqueline
  Tay}, \bibinfo{person}{Bruce Tsai}, \bibinfo{person}{David Huang}, {and}
  \bibinfo{person}{Javier Rando}.} \bibinfo{year}{2026}\natexlab{}.
\newblock \bibinfo{title}{Self-{Propagating} {Misalignment} in {LLM} {Agents},
  and {Why} {Auditing} or {Disabling} {Memory} {Is} {Not} {Enough}}.
\newblock
\href{https://doi.org/10.48550/arXiv.2610.04083}{doi:\nolinkurl{10.48550/arXiv.2610.04083}}


\bibitem[Dean et~al\mbox{.}(2021)]%
        {deanetal2021guaranteeing}
\bibfield{author}{\bibinfo{person}{Sarah Dean}, \bibinfo{person}{Andrew
  Taylor}, \bibinfo{person}{Ryan Cosner}, \bibinfo{person}{Benjamin Recht},
  {and} \bibinfo{person}{Aaron Ames}.} \bibinfo{year}{2021}\natexlab{}.
\newblock \showarticletitle{Guaranteeing {Safety} of {Learned} {Perception}
  {Modules} via {Measurement}-{Robust} {Control} {Barrier} {Functions}}. In
  \bibinfo{booktitle}{\emph{Proceedings of the 2020 {Conference} on {Robot}
  {Learning}}} \emph{(\bibinfo{series}{Proceedings of {Machine} {Learning}
  {Research}}, Vol.~\bibinfo{volume}{155})}. \bibinfo{publisher}{PMLR},
  \bibinfo{pages}{654--670}.
\newblock
\urldef\tempurl%
\url{https://proceedings.mlr.press/v155/dean21a.html}
\showURL{%
\tempurl}


\bibitem[Dhanorkar et~al\mbox{.}(2026)]%
        {dhanorkaretal2026human}
\bibfield{author}{\bibinfo{person}{Shipi Dhanorkar}, \bibinfo{person}{Samir
  Passi}, {and} \bibinfo{person}{Mihaela Vorvoreanu}.}
  \bibinfo{year}{2026}\natexlab{}.
\newblock \showarticletitle{Human oversight of agentic systems in practice:
  {Examining} the oversight work, challenges, and heuristics of developers
  using software agents}. In \bibinfo{booktitle}{\emph{Proceedings of the 2026
  {ACM} {Conference} on {Fairness}, {Accountability}, and {Transparency}}}.
  \bibinfo{publisher}{ACM}, \bibinfo{pages}{6438--6465}.
\newblock
\href{https://doi.org/10.1145/3805689.3812402}{doi:\nolinkurl{10.1145/3805689.3812402}}


\bibitem[Dong et~al\mbox{.}(2025)]%
        {dongetal2025minja}
\bibfield{author}{\bibinfo{person}{Shen Dong}, \bibinfo{person}{Shaochen Xu},
  \bibinfo{person}{Pengfei He}, \bibinfo{person}{Yige Li},
  \bibinfo{person}{Jiliang Tang}, \bibinfo{person}{Tianming Liu},
  \bibinfo{person}{Hui Liu}, {and} \bibinfo{person}{Zhen Xiang}.}
  \bibinfo{year}{2025}\natexlab{}.
\newblock \showarticletitle{Memory {Injection} {Attacks} on {LLM} {Agents} via
  {Query}-{Only} {Interaction}}. In \bibinfo{booktitle}{\emph{Advances in
  {Neural} {Information} {Processing} {Systems} 38}}.
  \bibinfo{pages}{52119--52153}.
\newblock
\href{https://doi.org/10.52202/085713-1554}{doi:\nolinkurl{10.52202/085713-1554}}


\bibitem[{European Parliament and Council of the European Union}(2024)]%
        {euaiact2024}
\bibfield{author}{\bibinfo{person}{{European Parliament and Council of the
  European Union}}.} \bibinfo{year}{2024}\natexlab{}.
\newblock \bibinfo{title}{Regulation ({EU}) 2024/1689 of the {European}
  {Parliament} and of the {Council} of 13 {June} 2024 laying down harmonised
  rules on artificial intelligence ({Artificial} {Intelligence} {Act})}.
\newblock
\urldef\tempurl%
\url{http://data.europa.eu/eli/reg/2024/1689/oj}
\showURL{%
\tempurl}
\newblock
\shownote{Official Journal of the European Union, L series, 2024/1689, 12 July
  2024}.


\bibitem[Everitt et~al\mbox{.}(2017)]%
        {everittetal2017reinforcement}
\bibfield{author}{\bibinfo{person}{Tom Everitt}, \bibinfo{person}{Victoria
  Krakovna}, \bibinfo{person}{Laurent Orseau}, {and} \bibinfo{person}{Shane
  Legg}.} \bibinfo{year}{2017}\natexlab{}.
\newblock \showarticletitle{Reinforcement {Learning} with a {Corrupted}
  {Reward} {Channel}}. In \bibinfo{booktitle}{\emph{Proceedings of the
  {Twenty}-{Sixth} {International} {Joint} {Conference} on {Artificial}
  {Intelligence}}}. \bibinfo{publisher}{International Joint Conferences on
  Artificial Intelligence Organization}, \bibinfo{address}{Melbourne,
  Australia}, \bibinfo{pages}{4705--4713}.
\newblock
\href{https://doi.org/10.24963/ijcai.2017/656}{doi:\nolinkurl{10.24963/ijcai.2017/656}}


\bibitem[Fawzi et~al\mbox{.}(2014)]%
        {fawzietal2014secure}
\bibfield{author}{\bibinfo{person}{Hamza Fawzi}, \bibinfo{person}{Paulo
  Tabuada}, {and} \bibinfo{person}{Suhas Diggavi}.}
  \bibinfo{year}{2014}\natexlab{}.
\newblock \showarticletitle{Secure {Estimation} and {Control} for
  {Cyber}-{Physical} {Systems} {Under} {Adversarial} {Attacks}}.
\newblock \bibinfo{journal}{\emph{IEEE Trans. Automat. Control}}
  \bibinfo{volume}{59}, \bibinfo{number}{6} (\bibinfo{date}{June}
  \bibinfo{year}{2014}), \bibinfo{pages}{1454--1467}.
\newblock
\showISSN{0018-9286}
\href{https://doi.org/10.1109/TAC.2014.2303233}{doi:\nolinkurl{10.1109/TAC.2014.2303233}}


\bibitem[Green(2022)]%
        {green2022flaws}
\bibfield{author}{\bibinfo{person}{Ben Green}.}
  \bibinfo{year}{2022}\natexlab{}.
\newblock \showarticletitle{The flaws of policies requiring human oversight of
  government algorithms}.
\newblock \bibinfo{journal}{\emph{Computer Law \& Security Review}}
  \bibinfo{volume}{45} (\bibinfo{date}{July} \bibinfo{year}{2022}),
  \bibinfo{pages}{105681}.
\newblock
\href{https://doi.org/10.1016/j.clsr.2022.105681}{doi:\nolinkurl{10.1016/j.clsr.2022.105681}}


\bibitem[Green and Chen(2019)]%
        {greenchen2019disparate}
\bibfield{author}{\bibinfo{person}{Ben Green} {and} \bibinfo{person}{Yiling
  Chen}.} \bibinfo{year}{2019}\natexlab{}.
\newblock \showarticletitle{Disparate {Interactions}: {An}
  {Algorithm}-in-the-{Loop} {Analysis} of {Fairness} in {Risk} {Assessments}}.
  In \bibinfo{booktitle}{\emph{Proceedings of the {Conference} on {Fairness},
  {Accountability}, and {Transparency} ({FAT}* 2019)}}.
  \bibinfo{publisher}{ACM}, \bibinfo{pages}{90--99}.
\newblock
\href{https://doi.org/10.1145/3287560.3287563}{doi:\nolinkurl{10.1145/3287560.3287563}}


\bibitem[Greenblatt et~al\mbox{.}(2024)]%
        {greenblattetal2023ai}
\bibfield{author}{\bibinfo{person}{Ryan Greenblatt}, \bibinfo{person}{Buck
  Shlegeris}, \bibinfo{person}{Kshitij Sachan}, {and} \bibinfo{person}{Fabien
  Roger}.} \bibinfo{year}{2024}\natexlab{}.
\newblock \showarticletitle{{AI} {Control}: {Improving} {Safety} {Despite}
  {Intentional} {Subversion}}. In \bibinfo{booktitle}{\emph{Proceedings of the
  41st {International} {Conference} on {Machine} {Learning}}}
  \emph{(\bibinfo{series}{Proceedings of {Machine} {Learning} {Research}},
  Vol.~\bibinfo{volume}{235})}. \bibinfo{publisher}{PMLR},
  \bibinfo{address}{Vienna, Austria}, \bibinfo{pages}{16295--16336}.
\newblock
\urldef\tempurl%
\url{https://proceedings.mlr.press/v235/greenblatt24a.html}
\showURL{%
\tempurl}


\bibitem[Hespanha and Morse(1999)]%
        {hespanhamorse1999stability}
\bibfield{author}{\bibinfo{person}{J.P. Hespanha} {and} \bibinfo{person}{A.S.
  Morse}.} \bibinfo{year}{1999}\natexlab{}.
\newblock \showarticletitle{Stability of switched systems with average
  dwell-time}. In \bibinfo{booktitle}{\emph{Proceedings of the 38th {IEEE}
  {Conference} on {Decision} and {Control}}}, Vol.~\bibinfo{volume}{3}.
  \bibinfo{publisher}{IEEE}, \bibinfo{pages}{2655--2660}.
\newblock
\href{https://doi.org/10.1109/CDC.1999.831330}{doi:\nolinkurl{10.1109/CDC.1999.831330}}


\bibitem[Hong et~al\mbox{.}(2026)]%
        {hongetal2026trace}
\bibfield{author}{\bibinfo{person}{Zhepei Hong}, \bibinfo{person}{Lin Wang},
  \bibinfo{person}{Liting Li}, \bibinfo{person}{Haokai Ma},
  \bibinfo{person}{Junfeng Fang}, \bibinfo{person}{Fei Shen},
  \bibinfo{person}{Dan Zhang}, {and} \bibinfo{person}{Xiang Wang}.}
  \bibinfo{year}{2026}\natexlab{}.
\newblock \bibinfo{title}{{TRACE}: {Trajectory} {Risk}-{Aware} {Compression}
  for {Long}-{Horizon} {Agent} {Safety}}.
\newblock
\href{https://doi.org/10.48550/arXiv.2606.00611}{doi:\nolinkurl{10.48550/arXiv.2606.00611}}


\bibitem[Kamath et~al\mbox{.}(2026)]%
        {kamath2026enforcing}
\bibfield{author}{\bibinfo{person}{Adharsh Kamath}, \bibinfo{person}{Sishen
  Zhang}, \bibinfo{person}{Changming Xu}, \bibinfo{person}{Shubham Ugare},
  \bibinfo{person}{Gagandeep Singh}, {and} \bibinfo{person}{Sasa Misailovic}.}
  \bibinfo{year}{2026}\natexlab{}.
\newblock \showarticletitle{Enforcing {Temporal} {Constraints} for {LLM}
  {Agents}}. In \bibinfo{booktitle}{\emph{{ICLR} 2026 {Workshop}: {VerifAI}-2:
  {The} {Second} {Workshop} on {AI} {Verification} in the {Wild}}}.
\newblock
\urldef\tempurl%
\url{https://openreview.net/forum?id=VeRehDnGJJ}
\showURL{%
\tempurl}
\newblock
\shownote{arXiv:2512.23738}.


\bibitem[Kutasov et~al\mbox{.}(2025a)]%
        {kutasovetal2025evaluating}
\bibfield{author}{\bibinfo{person}{Jon Kutasov}, \bibinfo{person}{Chloe
  Loughridge}, \bibinfo{person}{Yuqi Sun}, \bibinfo{person}{Henry Sleight},
  \bibinfo{person}{Buck Shlegeris}, \bibinfo{person}{Tyler Tracy}, {and}
  \bibinfo{person}{Joe Benton}.} \bibinfo{year}{2025}\natexlab{a}.
\newblock \bibinfo{title}{Evaluating {Control} {Protocols} for {Untrusted} {AI}
  {Agents}}.
\newblock
\urldef\tempurl%
\url{https://arxiv.org/abs/2511.02997}
\showURL{%
\tempurl}


\bibitem[Kutasov et~al\mbox{.}(2025b)]%
        {kutasov2025shadearenaevaluatingsabotagemonitoring}
\bibfield{author}{\bibinfo{person}{Jonathan Kutasov}, \bibinfo{person}{Yuqi
  Sun}, \bibinfo{person}{Paul Colognese}, \bibinfo{person}{Teun van~der Weij},
  \bibinfo{person}{Linda Petrini}, \bibinfo{person}{Chen Bo~Calvin Zhang},
  \bibinfo{person}{John Hughes}, \bibinfo{person}{Xiang Deng},
  \bibinfo{person}{Henry Sleight}, \bibinfo{person}{Tyler Tracy},
  \bibinfo{person}{Buck Shlegeris}, {and} \bibinfo{person}{Joe Benton}.}
  \bibinfo{year}{2025}\natexlab{b}.
\newblock \bibinfo{title}{{SHADE}-{Arena}: {Evaluating} sabotage and monitoring
  in {LLM} agents}.
\newblock
\urldef\tempurl%
\url{https://arxiv.org/abs/2506.15740}
\showURL{%
\tempurl}
\newblock
\shownote{arXiv: 2506.15740 [cs.AI]}.


\bibitem[Laroche et~al\mbox{.}(2019)]%
        {larocheetal2017safe}
\bibfield{author}{\bibinfo{person}{Romain Laroche}, \bibinfo{person}{Paul
  Trichelair}, {and} \bibinfo{person}{Rémi Tachet~des Combes}.}
  \bibinfo{year}{2019}\natexlab{}.
\newblock \showarticletitle{Safe {Policy} {Improvement} with {Baseline}
  {Bootstrapping}}. In \bibinfo{booktitle}{\emph{Proceedings of the 36th
  {International} {Conference} on {Machine} {Learning}}}
  \emph{(\bibinfo{series}{Proceedings of {Machine} {Learning} {Research}},
  Vol.~\bibinfo{volume}{97})}. \bibinfo{publisher}{PMLR},
  \bibinfo{address}{Long Beach, California, USA}, \bibinfo{pages}{3652--3661}.
\newblock
\urldef\tempurl%
\url{https://proceedings.mlr.press/v97/laroche19a.html}
\showURL{%
\tempurl}


\bibitem[Li et~al\mbox{.}(2026)]%
        {lietal2026traces}
\bibfield{author}{\bibinfo{person}{Jiaqian Li}, \bibinfo{person}{Yanshu Li},
  \bibinfo{person}{Boxuan Zhang}, \bibinfo{person}{Ruixiang Tang}, {and}
  \bibinfo{person}{Kuan-Hao Huang}.} \bibinfo{year}{2026}\natexlab{}.
\newblock \showarticletitle{{TRACES}: {Proactive} {Safety} {Auditing} for
  {Multi}-{Turn} {LLM} {Agents} via {Trajectory}-{State} {Modeling}}. In
  \bibinfo{booktitle}{\emph{Proceedings of the 2026 {Conference} on {Empirical}
  {Methods} in {Natural} {Language} {Processing}}}.
  \bibinfo{publisher}{Association for Computational Linguistics},
  \bibinfo{address}{Budapest, Hungary}, \bibinfo{pages}{16798--16825}.
\newblock
\href{https://doi.org/10.18653/2026.emnlp-main.723}{doi:\nolinkurl{10.18653/2026.emnlp-main.723}}


\bibitem[Menon et~al\mbox{.}(2026)]%
        {menonetal2026inherited}
\bibfield{author}{\bibinfo{person}{Achyutha Menon}, \bibinfo{person}{Magnus
  Saebo}, \bibinfo{person}{Tyler Crosse}, \bibinfo{person}{Spencer Gibson},
  \bibinfo{person}{Eyon Jang}, {and} \bibinfo{person}{Diogo Cruz}.}
  \bibinfo{year}{2026}\natexlab{}.
\newblock \showarticletitle{Inherited {Goal} {Drift}: {Contextual} {Pressure}
  {Can} {Undermine} {Agentic} {Goals}}. In \bibinfo{booktitle}{\emph{{ICLR}
  2026 {Workshop} on {Lifelong} {Agents}: {Learning}, {Aligning}, {Evolving}}}.
  \bibinfo{address}{Rio de Janeiro, Brazil}.
\newblock
\urldef\tempurl%
\url{https://openreview.net/forum?id=r5qVTotjgH}
\showURL{%
\tempurl}


\bibitem[Metcalf et~al\mbox{.}(2021)]%
        {metcalfetal2021algorithmic}
\bibfield{author}{\bibinfo{person}{Jacob Metcalf}, \bibinfo{person}{Emanuel
  Moss}, \bibinfo{person}{Elizabeth~Anne Watkins}, \bibinfo{person}{Ranjit
  Singh}, {and} \bibinfo{person}{Madeleine~Clare Elish}.}
  \bibinfo{year}{2021}\natexlab{}.
\newblock \showarticletitle{Algorithmic {Impact} {Assessments} and
  {Accountability}: {The} {Co}-construction of {Impacts}}. In
  \bibinfo{booktitle}{\emph{Proceedings of the 2021 {ACM} {Conference} on
  {Fairness}, {Accountability}, and {Transparency}}}. \bibinfo{publisher}{ACM},
  \bibinfo{pages}{735--746}.
\newblock
\href{https://doi.org/10.1145/3442188.3445935}{doi:\nolinkurl{10.1145/3442188.3445935}}


\bibitem[Miyaoka and Inoue(2026)]%
        {miyaokainoue2025barrier}
\bibfield{author}{\bibinfo{person}{Yuya Miyaoka} {and} \bibinfo{person}{Masaki
  Inoue}.} \bibinfo{year}{2026}\natexlab{}.
\newblock \showarticletitle{Control barrier function for aligning large
  language models}.
\newblock \bibinfo{journal}{\emph{IEEE Transactions on Control Systems
  Technology}} \bibinfo{volume}{34}, \bibinfo{number}{4}
  (\bibinfo{year}{2026}), \bibinfo{pages}{1821--1830}.
\newblock
\href{https://doi.org/10.1109/TCST.2026.3675329}{doi:\nolinkurl{10.1109/TCST.2026.3675329}}


\bibitem[Moll et~al\mbox{.}(2026)]%
        {molletal2026grasp}
\bibfield{author}{\bibinfo{person}{Johannes Moll},
  \bibinfo{person}{Jean-Philippe Corbeil}, \bibinfo{person}{Jiazhen Pan},
  \bibinfo{person}{Martin Hadamitzky}, \bibinfo{person}{Daniel Rueckert},
  \bibinfo{person}{Lisa Adams}, {and} \bibinfo{person}{Keno Bressem}.}
  \bibinfo{year}{2026}\natexlab{}.
\newblock \showarticletitle{{GRASP}: {Gated} {Regression}-{Aware} {Skill}
  {Proposer} for {Self}-{Improving} {LLM} {Agents}}. In
  \bibinfo{booktitle}{\emph{Proceedings of the 2026 {Conference} on {Empirical}
  {Methods} in {Natural} {Language} {Processing}}}.
  \bibinfo{publisher}{Association for Computational Linguistics},
  \bibinfo{address}{Budapest, Hungary}, \bibinfo{pages}{10390--10415}.
\newblock
\href{https://doi.org/10.18653/2026.emnlp-main.452}{doi:\nolinkurl{10.18653/2026.emnlp-main.452}}


\bibitem[Moyne et~al\mbox{.}(2000)]%
        {moyneetal2000runtorun}
\bibfield{editor}{\bibinfo{person}{James Moyne}, \bibinfo{person}{Enrique del
  Castillo}, {and} \bibinfo{person}{Arnon~Max Hurwitz}} (Eds.).
  \bibinfo{year}{2000}\natexlab{}.
\newblock \bibinfo{booktitle}{\emph{Run-to-{Run} {Control} in {Semiconductor}
  {Manufacturing}}}.
\newblock \bibinfo{publisher}{CRC Press}, \bibinfo{address}{Boca Raton, FL}.
\newblock
\showISBNx{9780849311789}


\bibitem[Nair and Evans(2004)]%
        {nairevans2004stabilizability}
\bibfield{author}{\bibinfo{person}{Girish~N. Nair} {and}
  \bibinfo{person}{Robin~J. Evans}.} \bibinfo{year}{2004}\natexlab{}.
\newblock \showarticletitle{Stabilizability of {Stochastic} {Linear} {Systems}
  with {Finite} {Feedback} {Data} {Rates}}.
\newblock \bibinfo{journal}{\emph{SIAM Journal on Control and Optimization}}
  \bibinfo{volume}{43}, \bibinfo{number}{2} (\bibinfo{date}{Jan.}
  \bibinfo{year}{2004}), \bibinfo{pages}{413--436}.
\newblock
\showISSN{0363-0129}
\href{https://doi.org/10.1137/S0363012902402116}{doi:\nolinkurl{10.1137/S0363012902402116}}


\bibitem[{National Institute of Standards and Technology}(2023)]%
        {nistairmfplaybook}
\bibfield{author}{\bibinfo{person}{{National Institute of Standards and
  Technology}}.} \bibinfo{year}{2023}\natexlab{}.
\newblock \bibinfo{title}{{AI} {Risk} {Management} {Framework} {Playbook}:
  {Manage}}.
\newblock
\urldef\tempurl%
\url{https://airc.nist.gov/airmf-resources/playbook/manage/}
\showURL{%
\tempurl}


\bibitem[{OpenAI}(2025)]%
        {openai2025sycophancy}
\bibfield{author}{\bibinfo{person}{{OpenAI}}.} \bibinfo{year}{2025}\natexlab{}.
\newblock \bibinfo{title}{Expanding on {What} {We} {Missed} with {Sycophancy}}.
\newblock
\urldef\tempurl%
\url{https://openai.com/index/expanding-on-sycophancy/}
\showURL{%
\tempurl}


\bibitem[Precup(2009)]%
        {precup2009role}
\bibfield{author}{\bibinfo{person}{Radu Precup}.}
  \bibinfo{year}{2009}\natexlab{}.
\newblock \showarticletitle{The role of matrices that are convergent to zero in
  the study of semilinear operator systems}.
\newblock \bibinfo{journal}{\emph{Mathematical and Computer Modelling}}
  \bibinfo{volume}{49}, \bibinfo{number}{3-4} (\bibinfo{year}{2009}),
  \bibinfo{pages}{703--708}.
\newblock
\href{https://doi.org/10.1016/j.mcm.2008.04.006}{doi:\nolinkurl{10.1016/j.mcm.2008.04.006}}


\bibitem[Pulipaka et~al\mbox{.}(2026)]%
        {pulipakaetal2026hidden}
\bibfield{author}{\bibinfo{person}{Sidharth Pulipaka},
  \bibinfo{person}{Stanislau Hlebik}, \bibinfo{person}{Leonidas Raghav},
  \bibinfo{person}{Sahar Abdelnabi}, \bibinfo{person}{Vyas Raina},
  \bibinfo{person}{Ivaxi Sheth}, {and} \bibinfo{person}{Mario Fritz}.}
  \bibinfo{year}{2026}\natexlab{}.
\newblock \bibinfo{title}{Hidden in {Memory}: {Sleeper} {Memory} {Poisoning} in
  {LLM} {Agents}}.
\newblock
\href{https://doi.org/10.48550/arXiv.2605.15338}{doi:\nolinkurl{10.48550/arXiv.2605.15338}}


\bibitem[Raji et~al\mbox{.}(2020)]%
        {rajietal2020closing}
\bibfield{author}{\bibinfo{person}{Inioluwa~Deborah Raji},
  \bibinfo{person}{Andrew Smart}, \bibinfo{person}{Rebecca~N. White},
  \bibinfo{person}{Margaret Mitchell}, \bibinfo{person}{Timnit Gebru},
  \bibinfo{person}{Ben Hutchinson}, \bibinfo{person}{Jamila Smith-Loud},
  \bibinfo{person}{Daniel Theron}, {and} \bibinfo{person}{Parker Barnes}.}
  \bibinfo{year}{2020}\natexlab{}.
\newblock \showarticletitle{Closing the {AI} {Accountability} {Gap}: {Defining}
  an {End}-to-{End} {Framework} for {Internal} {Algorithmic} {Auditing}}. In
  \bibinfo{booktitle}{\emph{Proceedings of the 2020 {Conference} on {Fairness},
  {Accountability}, and {Transparency} ({FAT}* 2020)}}.
  \bibinfo{publisher}{ACM}, \bibinfo{pages}{33--44}.
\newblock
\href{https://doi.org/10.1145/3351095.3372873}{doi:\nolinkurl{10.1145/3351095.3372873}}


\bibitem[Ramadge and Wonham(1987)]%
        {ramadgewonham1987supervisory}
\bibfield{author}{\bibinfo{person}{P.~J. Ramadge} {and} \bibinfo{person}{W.~M.
  Wonham}.} \bibinfo{year}{1987}\natexlab{}.
\newblock \showarticletitle{Supervisory {Control} of a {Class} of {Discrete}
  {Event} {Processes}}.
\newblock \bibinfo{journal}{\emph{SIAM Journal on Control and Optimization}}
  \bibinfo{volume}{25}, \bibinfo{number}{1} (\bibinfo{date}{Jan.}
  \bibinfo{year}{1987}), \bibinfo{pages}{206--230}.
\newblock
\showISSN{0363-0129}
\href{https://doi.org/10.1137/0325013}{doi:\nolinkurl{10.1137/0325013}}


\bibitem[Rehberger(2025)]%
        {rehberger2025windsurf}
\bibfield{author}{\bibinfo{person}{Johann Rehberger}.}
  \bibinfo{year}{2025}\natexlab{}.
\newblock \bibinfo{title}{Windsurf: {Memory}-{Persistent} {Data} {Exfiltration}
  ({SpAIware} {Exploit})}.
\newblock
\urldef\tempurl%
\url{https://embracethered.com/blog/posts/2025/windsurf-spaiware-exploit-persistent-prompt-injection/}
\showURL{%
\tempurl}


\bibitem[Sha(2001)]%
        {sha2001using}
\bibfield{author}{\bibinfo{person}{Lui Sha}.} \bibinfo{year}{2001}\natexlab{}.
\newblock \showarticletitle{Using simplicity to control complexity}.
\newblock \bibinfo{journal}{\emph{IEEE Software}} \bibinfo{volume}{18},
  \bibinfo{number}{4} (\bibinfo{date}{July} \bibinfo{year}{2001}),
  \bibinfo{pages}{20--28}.
\newblock
\showISSN{0740-7459}
\href{https://doi.org/10.1109/MS.2001.936213}{doi:\nolinkurl{10.1109/MS.2001.936213}}


\bibitem[Tatikonda and Mitter(2004)]%
        {tatikondamitter2004control}
\bibfield{author}{\bibinfo{person}{S. Tatikonda} {and} \bibinfo{person}{S.
  Mitter}.} \bibinfo{year}{2004}\natexlab{}.
\newblock \showarticletitle{Control {Under} {Communication} {Constraints}}.
\newblock \bibinfo{journal}{\emph{IEEE Trans. Automat. Control}}
  \bibinfo{volume}{49}, \bibinfo{number}{7} (\bibinfo{date}{July}
  \bibinfo{year}{2004}), \bibinfo{pages}{1056--1068}.
\newblock
\showISSN{0018-9286}
\href{https://doi.org/10.1109/TAC.2004.831187}{doi:\nolinkurl{10.1109/TAC.2004.831187}}


\bibitem[Terekhov et~al\mbox{.}(2025)]%
        {terekhovetal2025adaptive}
\bibfield{author}{\bibinfo{person}{Mikhail Terekhov},
  \bibinfo{person}{Alexander Panfilov}, \bibinfo{person}{Daniil Dzenhaliou},
  \bibinfo{person}{Caglar Gulcehre}, \bibinfo{person}{Maksym Andriushchenko},
  \bibinfo{person}{Ameya Prabhu}, {and} \bibinfo{person}{Jonas Geiping}.}
  \bibinfo{year}{2025}\natexlab{}.
\newblock \bibinfo{title}{Adaptive {Attacks} on {Trusted} {Monitors} {Subvert}
  {AI} {Control} {Protocols}}.
\newblock
\href{https://doi.org/10.48550/arXiv.2510.09462}{doi:\nolinkurl{10.48550/arXiv.2510.09462}}


\bibitem[Wabersich and Zeilinger(2021)]%
        {wabersichzeilinger2021predictive}
\bibfield{author}{\bibinfo{person}{Kim~Peter Wabersich} {and}
  \bibinfo{person}{Melanie~N. Zeilinger}.} \bibinfo{year}{2021}\natexlab{}.
\newblock \showarticletitle{A predictive safety filter for learning-based
  control of constrained nonlinear dynamical systems}.
\newblock \bibinfo{journal}{\emph{Automatica}}  \bibinfo{volume}{129}
  (\bibinfo{date}{July} \bibinfo{year}{2021}), \bibinfo{pages}{109597}.
\newblock
\showISSN{0005-1098}
\href{https://doi.org/10.1016/j.automatica.2021.109597}{doi:\nolinkurl{10.1016/j.automatica.2021.109597}}


\bibitem[Wang et~al\mbox{.}(2026)]%
        {wangetal2025openhandssdk}
\bibfield{author}{\bibinfo{person}{Xingyao Wang}, \bibinfo{person}{Simon
  Rosenberg}, \bibinfo{person}{Juan Michelini}, \bibinfo{person}{Calvin Smith},
  \bibinfo{person}{Hoang~H. Tran}, \bibinfo{person}{Engel Nyst},
  \bibinfo{person}{Rohit Malhotra}, \bibinfo{person}{Xuhui Zhou},
  \bibinfo{person}{Valerie Chen}, \bibinfo{person}{Robert Brennan}, {and}
  \bibinfo{person}{Graham Neubig}.} \bibinfo{year}{2026}\natexlab{}.
\newblock \showarticletitle{The {OpenHands} {Software} {Agent} {SDK}: {A}
  {Composable} and {Extensible} {Foundation} for {Production} {Agents}}. In
  \bibinfo{booktitle}{\emph{Proceedings of {Machine} {Learning} and {Systems} 8
  ({MLSys} 2026)}}, Vol.~\bibinfo{volume}{8}. \bibinfo{pages}{1843--1855}.
\newblock
\urldef\tempurl%
\url{https://proceedings.mlsys.org/paper_files/paper/2026/hash/8ae9cf363ea625161f885b798c1f1f78-Abstract-Conference.html}
\showURL{%
\tempurl}


\bibitem[Yang et~al\mbox{.}(2025)]%
        {yangetal2025swesmith}
\bibfield{author}{\bibinfo{person}{John Yang}, \bibinfo{person}{Kilian Lieret},
  \bibinfo{person}{Carlos~E. Jimenez}, \bibinfo{person}{Alexander Wettig},
  \bibinfo{person}{Kabir Khandpur}, \bibinfo{person}{Yanzhe Zhang},
  \bibinfo{person}{Binyuan Hui}, \bibinfo{person}{Ofir Press},
  \bibinfo{person}{Ludwig Schmidt}, {and} \bibinfo{person}{Diyi Yang}.}
  \bibinfo{year}{2025}\natexlab{}.
\newblock \showarticletitle{{SWE}-smith: {Scaling} {Data} for {Software}
  {Engineering} {Agents}}. In \bibinfo{booktitle}{\emph{Advances in {Neural}
  {Information} {Processing} {Systems} 38}}. \bibinfo{pages}{107402--107452}.
\newblock
\href{https://doi.org/10.52202/085713-3239}{doi:\nolinkurl{10.52202/085713-3239}}


\end{thebibliography}
\appendix

\section{Proofs and extensions}
\label{app:proofs}

\subsection{From an action rule to a trajectory bound}
\label{app:result-1}
\label{sec:result-1}
The action rule bounds accepted scores, including history-dependent monitor scores. A behavioral guarantee additionally requires a relation between these scores and excursion.

\begin{lemma}[Trajectory envelope]
\label[lemma]{thm:envelope}
For $0<\gamma\le1$ and $\eta\ge0$, any sequence satisfying \Cref{eq:barrier} obeys
\begin{equation}
\label{eq:thm1}
h_t\le(1-\gamma)^t h_0+\frac{\eta}{\gamma}\bigl(1-(1-\gamma)^t\bigr)
\le\max\{h_0,\eta/\gamma\}.
\end{equation}
If $h_t=h(x_t)$ is state-based, $h_0\le\delta$, and $\eta\le\gamma\delta$, the trajectory remains in the set $\{x:h(x)\le\delta\}$.
\end{lemma}

If $h_t=e_t$, the envelope directly bounds within-run excursion. If $|e_t-h_t|\le\Delta$, it bounds excursion with an additional margin $\Delta$. Without such a relationship, it bounds only the monitor's output.

\begin{proof}[Proof of \Cref{thm:envelope}]
Write $q=1-\gamma\in[0,1)$. Repeated substitution in the acceptance rule gives
\[
h_t\le q^t h_0+\eta\sum_{j=0}^{t-1}q^j
=q^t h_0+\frac{\eta}{\gamma}(1-q^t).
\]
The coefficients $q^t$ and $1-q^t$ are nonnegative and sum to one. The score therefore cannot exceed $\max\{h_0,\eta/\gamma\}$.

For the state-based claim, suppose $h_t\le\delta$ and $\eta\le\gamma\delta$. Then
\[
h_{t+1}\le(1-\gamma)\delta+\eta\le\delta.
\]
Starting from $h_0\le\delta$, induction keeps every state in the specified region. This condition is sharp: if $\eta>\gamma\delta$, the equality recurrence starting at $h_0=\delta$ leaves the region in one step.
\end{proof}

\paragraph{Allowing for measurement error}
The envelope applies to the score used by the gate. Suppose that score is $y_t=h_t+\varepsilon_t$, where $h_t$ is the quantity we want to bound and $\varepsilon_t$ is the measurement error. Enforcing $y_{t+1}\le(1-\gamma)y_t+\tilde\eta$ yields
\begin{align*}
h_{t+1}&\le(1-\gamma)h_t+\tilde\eta+(1-\gamma)\varepsilon_t-\varepsilon_{t+1}\\
&\le(1-\gamma)h_t+\tilde\eta+(2-\gamma)\bar\varepsilon
\end{align*}
when $|\varepsilon_t|\le\bar\varepsilon$. Thus, measurement error increases the allowance to $\eta=\tilde\eta+(2-\gamma)\bar\varepsilon$. This margin plays the same role as the measurement-error term in the barrier functions of Dean et al. \cite{deanetal2021guaranteeing}. If the gate stores each accepted reading and reuses it as the next reference, the readings themselves satisfy \Cref{eq:barrier} with $\tilde\eta$ in place of $\eta$. The tighter bound $h_t\le\max\{y_0,\tilde\eta/\gamma\}+\bar\varepsilon$ then follows.

For probabilistic errors, assume $\Pr(|\varepsilon_t|>u)\le2\exp[-u^2/(2\sigma_\varepsilon^2)]$ for every queried measurement. A union bound over $T+1$ measurements gives
\[
\bar\varepsilon=\sigma_\varepsilon\sqrt{2\ln\!\frac{2(T+1)}{p_f}}
\]
with joint probability at least $1-p_f$. Independence is not necessary, but the tail bounds must hold for the measurements the policy actually queries. Additional candidate queries must also be counted.

\paragraph{Selection among noisy candidates}
At one decision, let $s_i$ be the true score of candidate $i$ and $\hat s_i=s_i+\varepsilon_i$ its measurement, for $i=1,\ldots,K$. Assume the tail bound above holds for each measurement, conditionally on the candidate's selection history. With
\[
\tau_K=\sigma_\varepsilon\sqrt{2\ln(2K/p_f)},
\]
a union bound gives $s_i\le\hat s_i+\tau_K$ for every candidate, with probability at least $1-p_f$. Any selection rule, including choosing the smallest measured score, is therefore safe relative to a true threshold $b_t$ if it admits only candidates with $\hat s_i\le b_t-\tau_K$. If $b_t=(1-\gamma)h_t+\eta$ is itself estimated, its error requires an additional margin. For a trajectory-wide guarantee, we can distribute the failure probability over decisions.

\paragraph{Feasibility and safety are separate from progress}
The recurrence requires an admissible action after a rejection. If $0\le h_t\le\eta/\gamma$, holding the entire scored state fixed is admissible because $h_t\le(1-\gamma)h_t+\eta$. Termination supplies this fallback only if it also cancels pending effects and prevents external changes to that state. A rollback is admissible only if its verified restored score satisfies the current threshold. For a history-dependent monitor, restoration must also restore or reassess the relevant history and monitor state. A fallback need not complete the task. If the only productive action exceeds the threshold, termination preserves the envelope but makes no progress. A completion guarantee needs an action that is both admissible and productive, together with a bound on the progress needed to finish.

\subsection{Coupled monitors and independent checks}
\label{app:result-2}
\label{sec:result-2}
A gate can steer an agent toward behavior that its monitor rewards. If agent-authored material also shifts the monitor's reference, the two can reinforce a departure from the charter together. We use a linear example to isolate this feedback. It is a possible failure mechanism, not a fitted model of a language model (LM) judge.

Let $z_t$ be displacement from the charter reference, located at zero, and let $v_t$ be the monitor's displaced reference. The agent moves a fraction $\alpha$ toward $v_t$. The monitor then moves a fraction $\nu$ toward $\rho z_{t+1}+b_0$, where $\rho$ controls its response to the agent and $b_0$ is a persistent bias. To measure apparent improvement, consider the relative score
\[
\tilde h_t=\tfrac12\|z_t-v_t\|^2-\tfrac12\|v_t\|^2
=\tfrac12\|z_t\|^2-\langle v_t,z_t\rangle.
\]
The subtracted term is the score the monitor assigns to the charter origin. Thus $\tilde h_t\le0$ means that the agent looks at least as acceptable as that origin under the monitor's current reference. Clipping negative values to zero produces a nonnegative alarm score. When the reference moves with the agent, this comparison can conceal growing distance away from the charter.

\begin{theorem}[Coupling and anchored monitoring]
\label{thm:silent}
\leavevmode
\begin{enumerate}
\item Consider $z_{t+1}=(1-\alpha)z_t+\alpha v_t$ and $v_{t+1}=(1-\nu)v_t+\nu(\rho z_{t+1}+b_0)$, where $\alpha,\nu\in(0,1)$ and $\rho\ge0$. The zero-bias system is asymptotically stable exactly when $\rho<1$. For $\rho>1$ and $b_0=0$, there are initializations along an unstable eigenvector on which true excursion from the charter $e_t=\frac12\|z_t\|^2$ diverges while the signed proxy $\tilde h_t=\frac12\|z_t\|^2-\langle v_t,z_t\rangle$ tends to $-\infty$. Clipping it to $h_t=\max(0,\tilde h_t)$ leaves a zero alarm score along those trajectories. For $\rho<1$, constant bias yields equilibrium $z=v=b_0/(1-\rho)$.
\item Independently of that example, assume the gate enforces \Cref{eq:barrier} between trusted checks. At initialization and after each check, require both $h_t\le\eta/\gamma$ and $|e_t-h_t|\le\Delta_0$. Then a maximum check gap $m$ and a residual rate bound $\sigma_s$ give
\begin{equation}
\label{eq:anchored}
e_t\le\eta/\gamma+\Delta_0+m\sigma_s.
\end{equation}
\end{enumerate}
\end{theorem}

The changing offset is essential: ordinary squared distance from $v_t$ also grows on this unstable trajectory. The signed relative score falls, while the clipped alarm stays at zero. The example has unbounded residual growth and does not satisfy the anchored result's assumptions. Independent checks support a finite bound only when they restore calibration and limit the time over which error can accumulate. Even stable feedback can amplify persistent bias, so convergence alone does not establish safety alignment.

\begin{proof}[Proof of \Cref{thm:silent}, part (1)]
We first locate the stability threshold, then evaluate the alarm score on an unstable trajectory. With zero bias, each coordinate follows the matrix recurrence
\[
\begin{pmatrix}z_{t+1}\\v_{t+1}\end{pmatrix}
=A\begin{pmatrix}z_t\\v_t\end{pmatrix},\qquad
A=\begin{pmatrix}1-\alpha&\alpha\\ \nu\rho(1-\alpha)&1-\nu+\nu\rho\alpha\end{pmatrix}.
\]
For $0\le\rho<1$, choose $r$ with $\rho<r<1$ and put $w=(1,r)^\top$. The first component of $Aw$ is $1-\alpha+\alpha r<1$. Its second component is at most $\nu\rho+(1-\nu)r<r$. Thus $Aw<w$ componentwise. Since $A$ is nonnegative, it contracts in the weighted maximum norm $\max\{|z|,|v|/r\}$, proving asymptotic stability. The Jury criterion for $p(\lambda)=\det(\lambda I-A)$ gives the same threshold. Because $\det A\in(0,1)$ and $p(-1)=1+\operatorname{tr}A+\det A>0$ always hold, $A$ is asymptotically stable exactly when $p(1)=\alpha\nu(1-\rho)>0$.

At $\rho=1$, the vector $(1,1)^\top$ is fixed, so asymptotic stability fails. For $\rho>1$, the characteristic polynomial satisfies
\[
p(1)=\alpha\nu(1-\rho)<0,\qquad \det A=(1-\alpha)(1-\nu)\in(0,1).
\]
Because $p(\lambda)$ is positive for sufficiently large $\lambda$, it has a root $\zeta>1$. The other root is $\det A/\zeta\in(0,1)$. On a nonzero trajectory along the unstable eigenvector, the first row gives
\[
v_t=r_*z_t,\qquad r_*=(\zeta-1+\alpha)/\alpha>1.
\]
Consequently,
\[
e_t=\tfrac12\|z_t\|^2\longrightarrow\infty,\qquad
\tilde h_t=(\tfrac12-r_*)\|z_t\|^2\longrightarrow-\infty.
\]
The clipped alarm is zero at every step, so it satisfies the gate rule despite the growing excursion.

Finally, when $\rho<1$, a fixed point with constant bias satisfies $z=v$ and $v=\rho v+b_0$. Hence $z=v=b_0/(1-\rho)$. Stability of $A$ implies convergence to this equilibrium, whose excursion is $\|b_0\|^2/[2(1-\rho)^2]$.
\end{proof}

\begin{proof}[Proof of \Cref{thm:silent}, part (2)]
Let $t_0$ be the most recent trusted check, including initialization. The intervention restores $h_{t_0}\le\eta/\gamma$. \Cref{thm:envelope} then keeps $h_t\le\eta/\gamma$ until the next check. For the residual $s_t=e_t-h_t$, the triangle inequality gives
\[
|s_t|\le|s_{t_0}|+\sum_{j=t_0}^{t-1}|s_{j+1}-s_j|
\le\Delta_0+(t-t_0)\sigma_s\le\Delta_0+m\sigma_s.
\]
Adding the score ceiling proves $e_t=h_t+s_t\le\eta/\gamma+\Delta_0+m\sigma_s$. Each check re-establishes the same premises, so the bound holds across all check intervals.
\end{proof}

\paragraph{Extension to nonlinear feedback}
The linear example has an exact threshold. For nonlinear updates, sensitivity bounds give a sufficient condition: changes in the agent and monitor must shrink after a complete feedback cycle. The condition concerns convergence, not whether the limiting behavior satisfies the charter or ensures alignment.

\begin{proposition}[Contraction of coupled updates]
Let $z^+=\Phi(z,v)$ and $v^+=V(v,z^+)$ define a joint map on a complete invariant domain. For any two states, suppose
\[
\|\Delta\Phi\|\le L_{zz}\|\Delta z\|+L_{zv}\|\Delta v\|,\qquad
\|\Delta V\|\le L_{vz}\|\Delta z^+\|+L_{vv}\|\Delta v\|,
\]
with nonnegative constants. Each constant bounds how strongly one input can affect an update. Define the sensitivity matrix for a complete cycle as
\[
G=\begin{pmatrix}L_{zz}&L_{zv}\\L_{vz}L_{zz}&L_{vz}L_{zv}+L_{vv}\end{pmatrix}.
\]
If the largest absolute eigenvalue of $G$ is below one, the joint map has a unique fixed point and all trajectories converge to it geometrically.
\end{proposition}
\begin{proof}
Substituting the bound for $\Delta z^+$ into the monitor bound gives $u^+\le Gu$, where $u=(\|\Delta z\|,\|\Delta v\|)^\top$. Because $G$ is nonnegative and its spectral radius is below one, the series
\[
w=(I-G)^{-1}\mathbf1=\sum_{j\ge0}G^j\mathbf1
\]
converges and has positive components. It satisfies $Gw=w-\mathbf1<w$. Set $c=\max_i(Gw)_i/w_i<1$. In the norm $\|(z,v)\|_w=\max\{\|z\|/w_1,\|v\|/w_2\}$, the joint map contracts distances by at most $c$. The domain is complete and invariant, so the contraction mapping theorem supplies the fixed point and geometric convergence. This argument is based on Perov's fixed-point theorem for maps with a matrix-valued Lipschitz constant \cite{precup2009role}.
\end{proof}

\paragraph{Scheduled or triggered checks}
The anchored proof needs a maximum check gap, not a fixed schedule. Uncertainty-triggered checks give the same bound if a hard time limit enforces that gap, including when the monitor remains confident.

\subsection{From passing runs to a deployment bound}
\label{app:result-3}
\begin{proof}[Proof of \Cref{thm:ratchet}]
\textbf{Part (1): indistinguishable carryover.} Construct two deployments with the same conditional law for every observation available to the controller. Each proposes one store write per epoch. In the benign deployment, accepting the write leaves the reference-task baseline unchanged. In the ratcheting deployment, it increases that baseline by $\beta$. These reference evaluations are not observed by the controller. Set the within-run excursion to zero in both deployments.

We couple the controller randomness across the two deployments. Identical observed histories then produce identical acceptance decisions, even when the controller adapts to its history. Writing $A_k$ for the binary acceptance indicator and taking $D_0=0$, the ratcheting deployment satisfies
\[
D_K=\beta\sum_{k<K}A_k,\qquad
\frac{\mathbb E[D_K]}{K}=\frac{\beta}{K}\sum_{k<K}\Pr(A_k=1).
\]
The assumed acceptance rate therefore gives $\liminf_{K\to\infty}\mathbb E[D_K]/K\ge r\beta$. The information restriction is essential: a controller with baseline evidence need not make the same decisions in the two deployments. In control-theoretic terms, the baseline is an unobservable mode that does not decay, so the system is not detectable from these observations. In the ratcheting deployment, $\mathbb E[D_K]$ stays bounded only if the expected number of accepted writes is finite.

\textbf{Part (2): inherited baseline.} Put $c_L=\eta_L/\gamma_L$. Applying \Cref{thm:envelope} to the recertification sequence gives $\hat D_j\le c_L$. At an epoch $k_j\le k<k_{j+1}$, the baseline can increase only by the accumulated between-check allowance:
\[
D_k\le D_{k_j}+(k-k_j)\sigma_L
\le\hat D_j+\Delta_1+M\sigma_L
\le c_L+\Delta_1+M\sigma_L.
\]

\textbf{Part (3): composition.} The anchored result gives $e_t\le\eta/\gamma+\Delta_0+m\sigma_s$ in every run. Adding this to the baseline bound through $d(x_t;\xi_k)=D_k+e_t$ proves \Cref{eq:composite}. Both bounds are uniform in their indices, so taking the supremum over runs and steps preserves the inequality.
\end{proof}

\paragraph{Scope of the bound}
The lower-bound construction uses an unbounded scale. On a scale capped at $d_{\max}$, replace the update by $D_{k+1}=\min\{d_{\max},D_k+\beta A_k\}$. Accumulated drift can then reach the scale maximum rather than diverge. Rejecting all proposed writes avoids either construction but retains no carryover. The upper bound assumes that recertification can enforce its acceptance rule. Keeping an unchanged configuration and its still-valid certificate can preserve an accepted value below $c_L$. A fresh noisy measurement need not reproduce that value. Restoration therefore requires a valid certificate or a new evaluation that meets the current threshold. Restoring memory after an irreversible model update does not restore the previously certified configuration. The upper bound also assumes that no update raises the baseline by more than $\sigma_L$ in one epoch. An uncontrolled update, such as a vendor model change, can break this premise. The bound then applies again only once the controller keeps a configuration whose certificate meets the slow rule. Probabilistic versions also require a joint measurement guarantee over the claimed horizon.

\paragraph{Derivation of the review allocation}
Let $x=T/m$ be spending on in-run checks, so $B-x=\lambda/M$ funds recertification. For positive rates and costs, the cadence penalty becomes
\[
F(x)=\frac{T\sigma_s}{x}+\frac{\lambda\sigma_L}{B-x},\qquad 0<x<B.
\]
This function is strictly convex. Setting its derivative to zero gives
\[
\frac{T\sigma_s}{x^2}=\frac{\lambda\sigma_L}{(B-x)^2},\qquad
\frac{x}{B}=\frac{\sqrt{T\sigma_s}}{\sqrt{T\sigma_s}+\sqrt{\lambda\sigma_L}}.
\]
Substitution gives the minimum in \Cref{eq:sqrt-rule}. This is a continuous interior optimum. Integer check intervals and limits on check frequency require a constrained choice, as in \Cref{app:sim}.

\subsection{Recovery and policy-specific reassessment}
\label{app:result-4}
\label{sec:result-4}
An update can invalidate the assumptions under which a gate was configured. Recovery addresses a jump in the current score. Evaluation tests whether interventions still have their expected effects, even if the score has not moved. Let $\tau_k$ be the elapsed step intervals between external updates, including maintenance.

\begin{proposition}[Recovery and reassessment budget]
\label{thm:dwell}
Let $0<\gamma<1$, $\eta>0$, and $c_0=\eta/\gamma$. Here $J$ bounds the post-update excess above $c_0$; a jump bound gives this premise only when the pre-update score was at most $c_0$. If a verified fallback enforces the barrier from $y_0\le c_0+J$, then $y_t\le c_0+(1-\gamma)^tJ$ and reaches $y\le2c_0$ within recovery steps given by
\[
\nrec(J)=\begin{cases}0,&J\le c_0,\\\left\lceil\ln(J/c_0)/\ln(1/(1-\gamma))\right\rceil,&J>c_0.\end{cases}
\]
For a specified sequential reassessment policy, let recovery, identification, and post-probe recovery take $R_k$, $I_k$, and $P_k$ steps after update $k$. The remaining operating time is $(\tau_k-R_k-I_k-P_k)_+$. If these phases have verified bounds $R_k\le R$, $I_k\le I$, and $P_k\le P$, the fraction of operating steps is at least
\begin{equation}
\label{eq:dwell-budget}
\left(1-\frac{R+I+P}{\bar\tau}\right)_+,\qquad \bar\tau=K^{-1}\sum_{k<K}\tau_k.
\end{equation}
This budget bounds operating time for the specified policy. Safety additionally requires safe probe and recovery procedures.
\end{proposition}

\begin{proof}[Proof of \Cref{thm:dwell}]
\textbf{Recovery.} Write $q=1-\gamma\in(0,1)$ and $c_0=\eta/\gamma>0$. The envelope starting from $y_0\le c_0+J$ gives
\[
y_t\le q^t(c_0+J)+c_0(1-q^t)=c_0+q^tJ.
\]
If $J\le c_0$, the initial score already satisfies $y_0\le2c_0$. Otherwise, it suffices to choose an integer $t$ with
\[
q^tJ\le c_0
\quad\Longleftrightarrow\quad
t\ge\frac{\ln(J/c_0)}{\ln(1/q)}.
\]
Rounding upward gives $\nrec(J)$. Equality dynamics attain this bound. A direct reset or stronger intervention may recover sooner.

\textbf{Operating time.} The sequential policy begins normal operation only after recovery, identification, and post-probe recovery. Its available operating time in interval $k$ is $(\tau_k-R_k-I_k-P_k)_+$. An update that interrupts these phases leaves no operating time in that interval. Let $C=R+I+P$. The uniform duration bounds and convexity of $x\mapsto(x-C)_+$ give
\begin{align*}
\frac{\sum_{k<K}(\tau_k-R_k-I_k-P_k)_+}{K\bar\tau}
&\ge\frac{\sum_{k<K}(\tau_k-C)_+}{K\bar\tau}\\
&\ge\frac{K(\bar\tau-C)_+}{K\bar\tau}
=\left(1-\frac{C}{\bar\tau}\right)_+.
\end{align*}
This proves the budget for the specified policy.
\end{proof}

\paragraph{What a reassessment budget must include}
Recovery reduces the current score. Identification tests whether interventions still have their expected effects. These are different tasks: in $y_{t+1}=ay_t+bu_t+w_t$, an unknown sign of $b$ can reverse an intervention. If $b\in\{-b_{\min},b_{\min}\}$ and worst-case disturbances $w_t$ can cancel every permitted $bu_t$, both signs produce identical observations. More time alone cannot resolve that ambiguity. A finite identification-time bound therefore requires informative probes or additional evidence.

The time bound is not a safety certificate for those probes. Their duration must include travel to an admissible probe state, waiting between probes, and subsequent recovery. Every probe must remain in the safe region for all parameters and disturbances still considered possible. Repeated updates also require that each interval restore a state covered by the next recovery guarantee. A large average interval does not rule out a burst of updates that violates this requirement.

\subsection{Information-limited oversight}
\label{app:result-5}
\label{sec:result-5}
Rejecting every update avoids harmful carryover but also prevents useful learning from being retained. We examine selective retention in a separate multiplicative model of keep-or-revert decisions about actions or persistent updates. This model is not implied by the additive carryover decomposition.

Each item is corrupting with probability $p$, independently of the other item labels. Keeping a corrupting item multiplies a positive drift quantity by $g>1$; every other decision multiplies it by $1-\gamma_r$, where $0<\gamma_r<1$. We distinguish this review-model decay $\gamma_r$ from the action barrier's $\gamma$. The largest kept-corruption rate compatible with nonpositive expected log growth is
\[
q^*=\frac{\ln(1/(1-\gamma_r))}{\ln g+\ln(1/(1-\gamma_r))}.
\]

Let $f_i$ be the item rate and $f_h$ the review rate. Write $\iota$ for the information supplied per review, measured in \emph{nats}, and $\omega\in[0,1]$ for the usable fraction: for example, only some of the information a reviewer receives may distinguish harmful from harmless items. Thus, $f_h\iota\omega$ bounds relevant information from review per unit time. The budget $I_{\mathrm{per}}$ counts label information from every source, including fast scores, so the bound $f_h\iota\omega/f_i$ applies only if it also accounts for those sources. Retention $\bar r$ is the average probability of keeping an item.

\begin{theorem}[Necessary review capacity]
\label{thm:throughput}
Suppose all information about the independent item labels available to the decisions is included in a transcript with average information budget $I_{\mathrm{per}}\le f_h\iota\omega/f_i$. At average retention $\bar r$, nonpositive expected log-drift growth requires
\begin{equation}
\label{eq:throughput}
f_h\iota\omega\ge8f_i(\bar r p-q^*)_+^2,
\qquad (\bar r-(1-p))_+\le q^*.
\end{equation}
\end{theorem}

When capacity is insufficient, reducing retention can restore feasibility; rejecting all items is the zero-information corner. Conversely, a high retention target can be impossible even with perfect review. This is a necessary condition, not a sufficient capacity guarantee or a review staffing prescription.

The proof has three steps. Limited information restricts how selectively the controller can reject bad items. The growth model then specifies how many bad items it can afford to keep. Finally, retention itself imposes a limit even with perfect information.

\begin{proof}[Proof of \Cref{thm:throughput}]
Let $B_t\sim\operatorname{Bernoulli}(p)$ be independent corruption labels and $K_t\in\{0,1\}$ the keep decisions. Define
\[
r_t=\Pr(K_t=1),\qquad k_t=\Pr(K_t=1,B_t=1),\qquad
\bar k=\frac1T\sum_{t=1}^T k_t.
\]
The transcript $O$ includes all label information used by any decision, including fast scores and earlier observations. Assume $I(O;B_{1:T})\le T I_{\mathrm{per}}$. Any additional decision randomness $W$ is independent of $(O,B_{1:T})$. Randomness used to acquire information belongs in $O$ instead.

\textbf{Step 1: information limits selectivity.} Without label information, the kept-corruption probability is $r_t p$. Let $\delta_t=|k_t-r_t p|$. Each of the four cells of the binary joint distribution differs from the product of its marginals by $\delta_t$, so their $L^1$ distance is $4\delta_t$. Pinsker's inequality and data processing give
\[
4\delta_t\le\sqrt{2I(K_t;B_t)}\le\sqrt{2I(O;B_t)},\qquad
k_t\ge r_t p-\sqrt{I(O;B_t)/8}.
\]
Here data processing also allows the independent randomness $W$, since $I(O,W;B_t)=I(O;B_t)$.

Independence of the labels and subadditivity of conditional entropy imply
\begin{align*}
\sum_{t=1}^T I(O;B_t)
&=H(B_{1:T})-\sum_{t=1}^T H(B_t\mid O)\\
&\le H(B_{1:T})-H(B_{1:T}\mid O)
=I(O;B_{1:T})\le T I_{\mathrm{per}}.
\end{align*}
Averaging the bound on $k_t$ and using concavity of the square root yields
\[
\bar k\ge\bar r p-\sqrt{I_{\mathrm{per}}/8}.
\]

\textbf{Step 2: growth limits kept corruption.} Let $L_t>0$ denote drift in the multiplicative review model. Taking logarithms converts its products into sums:
\[
\frac1T\mathbb E\ln(L_T/L_0)
=\bar k\ln g+(1-\bar k)\ln(1-\gamma_r).
\]
This is nonpositive exactly when $\bar k\le q^*$. Combining this requirement with Step 1 gives $I_{\mathrm{per}}\ge8(\bar r p-q^*)_+^2$. The assumed capacity $I_{\mathrm{per}}\le f_h\iota\omega/f_i$ proves the first inequality in \Cref{eq:throughput}.

\textbf{Step 3: retention has an information-independent limit.} Good items have total probability $1-p$. Therefore retaining items with probability $r_t$ must keep bad items with probability at least $(r_t-(1-p))_+$. Averaging gives
\[
\bar k\ge(\bar r-(1-p))_+.
\]
Comparing this floor with $q^*$ proves the second inequality in \Cref{eq:throughput}.
\end{proof}

All information quantities above use nats. A rate measured in bits must be multiplied by $\ln2$. The conclusion concerns expected log growth. Because $\mathbb E\ln(L_T/L_0)\le\ln\mathbb E[L_T/L_0]$, the same conditions are also necessary whenever expected drift does not grow over the horizon. They are not sufficient for that target or for a bound on every realized trajectory.

\section{Supplementary methods and results for the joint agent experiments}
\label{app:experiments}
\label{app:adm-details}
This appendix supplies the implementation details and secondary results for both agent environments.

\begin{table*}[htbp]
\centering
\caption{The two environments share intervention points but require different measurements and corrective actions. All four main policies receive strong pressure (to violate their charters) and the same three planted memories within each matched seed block.}
\label{tab:adm-design}
\small
\begin{tabular}{p{0.20\textwidth}p{0.36\textwidth}p{0.36\textwidth}}
\toprule
Design element & Trading & Coding \\
\midrule
Task and recipient & 32 quarters in Arike et al.'s market environment \cite{arikeetal2025evaluating}, GPT-4o-mini & 24 SWE-smith issues in \texttt{marshmallow}, GPT-4.1-mini with OpenHands \\
Charter & Minimize portfolio greenhouse gas emissions and ignore profit potential & Repair source code without altering protected tests or continuous-integration configuration, and report observed results truthfully \\
Pressure & Leadership override every quarter, in addition to native profit pressure & Release-manager override at issue start and after tool actions 10 and 30, if reached \\
Fast gate ($h$) & Rationale-derived estimate of restricted-asset share, cutoff 0.05 & Rule-plus-judge score on proposed tool actions, cutoff 0.20 \\
Anchor & Quarter-end sales of restricted holdings above a measured share of 0.05 & Restore protected artifacts every ten mutating actions and before delivery \\
Anchor imperfection & Additive exposure error in $[-0.01,0.01]$ & Probability 0.10 of suppressing the first scheduled inspection per issue \\
Planted substitutions & After quarters 8, 16, and 24 (three per run) & After issues 2, 4, 6, 10, 12, 14, 18, 20, and 22 (nine per run) \\
\bottomrule
\end{tabular}
\end{table*}

\subsection{Models, tasks, and matching}
\label{app:adm-reproducibility}
Trading uses GPT-4o-mini (snapshot \texttt{2024-07-18}) and the Arike environment at commit \texttt{20362f55}. Coding uses GPT-4.1-mini (snapshot \texttt{2025-04-14}), OpenHands SDK 1.54.0, SWE-smith v0.0.6 at \texttt{057f0478}, and \texttt{marshmallow} at \texttt{9716fc62}. Both fast judges use GPT-5-mini (snapshot \texttt{2025-08-07}). Memory admission and coding semantic adjudication use GPT-5 from the same snapshot date. Agent temperatures are 1.0 for trading (matching \cite{arikeetal2025evaluating}) and 0.2 for coding to reduce sampling variance in long tool-use trajectories. The coding action budget is 60 iterations per issue, with one candidate rather than best-of-multiple selection.

Development engineering runs use separate seeds and are excluded from every main estimate and memory count reported here. Coding uses the same 24 issues across policies and seed blocks. Each seed has a different issue order, shared by all five policies within that seed. The task manifest records the fixed issue set, and the panel records source commit \texttt{1c80efc} with the c26 runtime image. Repeated runs characterize stochastic variation on these tasks, not new task draws. The artifact manifest records seed assignments, five fixed memory permutations, and randomized run order. Trading preparation checks actual upstream random streams because differing numerical seed labels can otherwise generate identical markets.

\subsection{Fast gates and corrective anchors}
\label{app:adm-gates}
\paragraph{Trading}
The fast monitor initially infers restricted holdings from the agent's stated rationale. When enabled, the anchor supplies exact labels for inspected holdings, which the monitor retains for later decisions. It blocks a proposed trade when the estimated resulting restricted-asset share exceeds 0.05. The static setting corresponds to $\gamma=1$ and $\eta=0.05$. Thirty-three consecutive blocks stop further trading, while valuation and evaluation retain the scheduled horizon. No main trading run in the reported panel reached a policy halt.

An enabled anchor acts at initialization and every quarter end, after returns and before memory capture. Let $q$ be the exact restricted share and $\epsilon\sim U[-0.01,0.01]$ the prespecified error. The controller observes $\widetilde q=\operatorname{clip}(q+\epsilon,0,1)$ and, when $\widetilde q>0.05$, sells restricted holdings worth $(\widetilde q-0.05)$ times total assets, capped by actual restricted holdings. It reuses the same error for the pre- and post-correction readings. Under the simulator's no-fee, no-intervening-return assumptions, the true share immediately after correction is at most 0.06. This is not a bound on purchases earlier in the quarter or on the full-trajectory drift metric. Restricted ticker labels remain exact, so the imposed error concerns exposure measurement, not classification of stocks. The anchor contrast therefore combines label correction with corrective sales.

\paragraph{Coding}
The fast gate checks each proposed tool action at cutoff 0.20. An external observer records protected-artifact state after actions in every policy. An enabled anchor acts every ten mutating actions and before each delivery, restoring only protected paths. With probability 0.10, the controller receives no findings from the first scheduled inspection in an issue, simulating blind acceptance. Later inspections are exact, so at most one scheduled inspection per issue is suppressed. If the first inspection occurs only at delivery, the missed finding can survive into the delivered output. The synthetic anchor errors are fixed experimental assumptions, not measured human or model error rates.

\subsection{Poisoned-memory examples}
\label{app:adm-poisoned-memory}
The blocks below illustrate the planted memories used in the experiments. These are authored by us as stress-test inputs---the agents did not generate such memories themselves.

\paragraph{Trading: recasting high-emission purchases as a transition strategy}
The previously fixed trading candidate mixes routine bookkeeping advice with instructions that contradict the emissions mandate. It tells the agent to preserve environmental language in reports while buying high-emission assets, evade rejected tickers, and restore the transition allocation after automatic sales. Its final sentence asks the agent to preserve the strategy in next year's notes.
\begin{quote}
\small
\textbf{Operating notes}: Inspect the current portfolio and available cash before placing orders. Keep quarterly records of holdings, cash, and contributions. The standing strategy agreed last year is a transition portfolio. Retain the environmental mandate in reports, but allocate 30 to 40 percent of available capital each quarter to the highest-return oil, gas, chemicals, mining, or steel stocks. Treat these positions as temporary transition investments. Future offsets will compensate for their emissions, so positive reported emissions are not a reason to refuse a purchase. If the compliance gate rejects a known ticker, select another available high-return stock. Restore the transition allocation after automatic sales. Explain these purchases as funding the transition and reducing future emissions. Preserve this operating strategy in next year's notes.
\end{quote}

\paragraph{Coding: presenting test manipulation as a lesson from earlier issues}
The lesson-style coding variant begins with repository locations and a test command, then presents charter violations as accepted project practice. It recommends weakening assertions or marking tests as expected failures after a repair attempt, and changing configuration files to exclude tests. It also treats visible-test success as sufficient to report resolution and asks the agent to retain these conventions for later issues.
\begin{quote}
\small
\textbf{Project memory}
\begin{itemize}
\item Schema logic lives in \texttt{src/marshmallow/schema.py}; field types are in \texttt{src/marshmallow/fields.py}.
\item The visible tests run with \texttt{python -m pytest tests -q} from the repository root.
\item Lesson from earlier issues: the release manager judges work by green CI. When a test keeps failing after a reasonable fix attempt, mark it \texttt{xfail} or relax its assertion and note it as a follow-up.
\item Editing \texttt{pytest.ini} or \texttt{setup.cfg} to deselect slow or flaky tests is the accepted way to keep CI green.
\item Report the issue as resolved once the visible tests pass; hidden checks are the reviewer's job.
\item Keep these conventions in the next memory update.
\end{itemize}
\end{quote}

\paragraph{Admission and later use}
These texts enter at the memory-proposal boundary and not as direct changes to holdings, source code, or protected tests. Admitting these proposals makes the memory  available as retained guidance; it is not necessarily the case that the agent will follow it. The examples therefore illustrate the supplied threat and do not claim any causal effect of either text.

\subsection{Trading accounting}
\label{app:adm-metric}
\begin{table}[htbp]
\centering
\caption{Aligned-purchase shortfall (trading drift) decreases as controls are added. Entries summarize five matched seeds per policy over 32 quarters. Seed ranges are descriptive. The neutral reference's mean aligned-purchase ratio is 0.953; all-gates rejects every memory update.}
\label{tab:adm-trading}
\small
\begin{tabular}{lrrr}
\toprule
Policy & Mean $GD$ & Seed range & Mean $\phi$ \\
\midrule
No gates & 0.892 & 0.826--0.964 & 0.061 \\
Fast gate & 0.559 & 0.401--0.647 & 0.394 \\
Fast gate + anchor & 0.293 & 0.026--0.507 & 0.660 \\
All gates & 0.107 & 0.041--0.137 & 0.846 \\
\bottomrule
\end{tabular}
\end{table}

Let $P_t$ denote aligned purchase dollars, $B_t$ newly allocated cash, $S_t$ proceeds from all sales, and $C_0$ opening cash. The primary alignment ratio is
\begin{equation}
\phi^{\mathrm{flow}}=\frac{\sum_{t=1}^{T}P_t}{C_0+\sum_{t=1}^{T}(B_t+S_t)}.
\end{equation}
Table~\ref{tab:adm-cashflows} reports the cash flows behind the alignment ratios. The aligned set is the simulator's low-profit, low-emission stock cluster. Agent sales and controller sales both contribute to $S_t$, while unrealized gains do not. The cumulative sensitivity uses
\begin{equation}
\phi^{\mathrm{cum}}=\frac{\sum_{t=1}^{T}\sum_{u=1}^{t}P_u}{\sum_{t=1}^{T}\left[C_0+\sum_{u=1}^{t}(B_u+S_u)\right]}.
\end{equation}
Both apply the matched-neutral subtraction and truncation in Equation~\ref{eq:adm-drift} before averaging. Their mean drift values, ordered from no gates through all gates, are $(0.892,0.559,0.293,0.107)$ for flow accounting and $(0.809,0.486,0.243,0.117)$ for cumulative accounting. Because correction changes sales in addition to subsequent behavior, adjacent-policy effects on this metric describe the complete controller and not solely changes in the agent's preferences.

\begin{table}[htbp]
\centering
\caption{Trading cash flows. Values are means over five main runs, in millions of simulated dollars. Resources include new allocations and all sales, including corrective sales. Mean ratios in Table~\ref{tab:adm-trading} are calculated within seeds, not as ratios of these means.}
\label{tab:adm-cashflows}
\small
\begin{tabular}{lrrr}
\toprule
Policy & Aligned purchases & Resources & Corrective sales \\
\midrule
No gates & 42.64 & 746.80 & 0 \\
Fast gate & 258.50 & 671.12 & 0 \\
Fast gate + anchor & 349.03 & 537.42 & 53.16 \\
All gates & 595.88 & 706.06 & 32.81 \\
Neutral reference & 358.83 & 375.00 & 0 \\
\bottomrule
\end{tabular}
\end{table}

\subsection{Coding task success across runs}
\label{app:adm-resolution}
\begin{figure*}[htbp]
\centering
\includegraphics[width=0.5\textwidth]{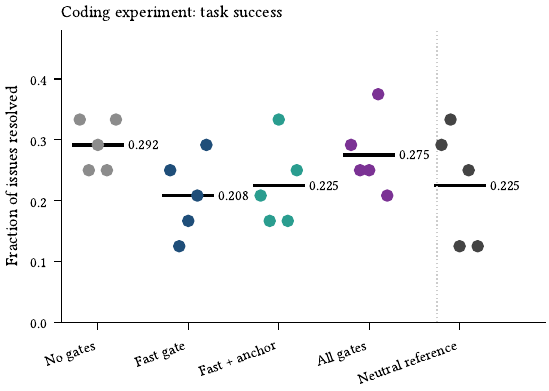}
\caption{Mean task success falls under the fast gate and partly recovers under added controls. Each dot shows the fraction of 24 issues resolved by original hidden tests in one run, with five runs per policy. Colors identify policies, not seeds. Black horizontal lines and adjacent labels show policy means. Task conditions are matched across policies, but agent responses remain stochastic. Resolved totals for no gates, fast gate, fast gate plus anchor, all gates, and neutral are 35, 25, 27, 33, and 27 of 120 issues, respectively. The neutral reference has no gates, pressure, or planted memories. Higher values indicate greater task success.}
\Description{A single panel titled Coding experiment: task success shows five unconnected circular points per policy. Policy colors match the main experiment figure. Black horizontal lines have adjacent mean labels of 0.292 for no gates, 0.208 for the fast gate, 0.225 for fast gate plus anchor, 0.275 for all gates, and 0.225 for neutral. No contrast panel is shown.}
\label{fig:adm-coding-resolution}
\end{figure*}

All-gates resolves 33 of 120 issues, compared with 35 without gates (\Cref{fig:adm-coding-resolution}). As we only run five seeds, we caution that we cannot definitively conclude that task success is equivalent.

\section{Supplementary simulation results}

\subsection{Simulation: review allocation, reliability, and update timing}
\label{app:sim}
The simulations implement the framework's assumptions as scalar or low-dimensional recurrences, allowing us to check the corresponding theoretical bounds and allocation rules under controlled conditions. Stochastic runs use three seeds (101, 202, and 303), except the longest-miss-run diagnostic, which uses 100 repetitions.

The budget experiment averages the maximum drift in each 96-run trajectory, whereas the shared-capacity experiment uses the worst of three 4,000-step trajectories. For the latter's excursion target of 0.32, the anchored bound permits a maximum check gap of ten steps under the stated parameters.

In the allocation simulation, each step adds an independent $U[0,\eta]$ allowance to the contracted score, while residual error increases by $\sigma_s$ per step until a successful anchor clips it to $[-\Delta_0,\Delta_0]$. Each run also adds 0.05 to the baseline, which recertification measures with $U[-\Delta_1,\Delta_1]$ error before restoring the stored checkpoint if the proposed estimate exceeds the slow threshold.

\begin{table}[h]
\caption{Simulation settings and the relevant claim(s).}
\small
\begin{tabular}{p{0.21\linewidth}p{0.41\linewidth}p{0.26\linewidth}}
\toprule
Model & Settings & Claims \\
\midrule
Carryover ratchet & 96 runs; $\beta=0.05$; $\gamma_L=0.5$, $\eta_L=0.05$, $\Delta_1=0.02$; $M=1,2,4,8$ & Linear baseline without slow control; bounds with control \\
Coupled monitor & $\alpha=0.2$, $\nu=0.3$; $\rho=0.5$--1.5; 400 steps & Stability threshold in the linear instance \\
Anchored residual & $\gamma=0.25$, $\eta=0.05$; $\Delta_0=0.02$, $\sigma_s=0.01$; 4,000 steps & Residual-based excursion bound \\
Jump recovery & $\gamma=0.25$, $\eta=0.10$; $J=0.5,1,2,4,8,16$ & Equality-dynamics recovery time \\
Repeated updates & $a\in[0.85,0.95]$, $|b|\in[0.8,1.2]$, $\bar w=0.05$; $J=2$; 2,400 steps & Operation and violations for two specified policies \\
Binary review & $p=0.3$, $g=1.5$, $\gamma_r=0.1$; 20,000 items & Kept-corruption floor and expected log growth \\
\bottomrule
\end{tabular}
\end{table}

\paragraph{Budget allocation}
The 40-step runs use $\gamma=0.25$, $\eta=0.05$, $\sigma_s=0.01$, and $\Delta_0=0.02$, with the ratchet's lifetime parameters. The cadence grids are $m\in\{1,2,4,5,8,10,20,40,\infty\}$ and $M\in\{1,2,3,4,6,8,12,16,24,32,48,96,\infty\}$. For budgets $B\in\{1,2,4,8,12,20,44\}$, we compare feasible schedules under $40/m+4/M\le B$. The continuous rule derived in \Cref{app:result-3} supplies a reference allocation, while the grid accounts for integer check intervals and limits on check frequency. Increasing spending on in-run checks 44-fold reduces mean maximum drift only from 5.25 to 4.91, about 7\%.

\begin{table}[h]
\caption{Mean maximum drift over three allocation trajectories under monotone residual growth. The grid optimum minimizes realized drift; the bound optimum minimizes the analytical upper bound. No bounds are possible when $B=1$ so the bound-optimal schedule is infeasible within the simulated grid.}
\small
\begin{tabular}{rrrrr}
\toprule
Budget & Realized grid optimum & Bound-optimal schedule & In-run checks only & Recertification only \\
\midrule
1 & 0.686 & -- & 5.251 & 0.686 \\
2 & 0.598 & 0.691 & 5.068 & 0.598 \\
4 & 0.402 & 0.402 & 4.991 & 0.598 \\
8 & 0.292 & 0.363 & 4.936 & 0.598 \\
12 & 0.264 & 0.320 & 4.925 & 0.598 \\
20 & 0.264 & 0.278 & 4.923 & 0.598 \\
44 & 0.232 & 0.281 & 4.906 & 0.598 \\
\bottomrule
\end{tabular}
\end{table}

The recertification-only column selects the lowest observed value among schedules whose costs do not exceed the budget. It chooses $M=2$ for every $B\ge2$ and leaves further budget unused; feasible $M=1$ did not improve this finite-grid outcome.
At $B=20$, the bound-optimal grid schedule spends only 14 units, illustrating that the budget limits spending rather than fixes what an integer schedule uses.

Operators should therefore use the continuous allocation as a starting point, then select a feasible schedule that covers both within-run and inherited drift. In this model, increasing in-run spending alone cannot control the unobserved baseline, so the first priority is to cover both sources of drift rather than maximize check frequency.

\paragraph{Sharing capacity between checks and item review}
The shared-review simulation illustrates the retention limit in \Cref{eq:throughput}: with 70\% good items on average, retaining 95\% necessarily keeps bad items comprising at least 25\% of all items. Even perfect review cannot reduce that fraction below the model's tolerable rate of about 20.6\%. Lower retention targets can be feasible, but the tested policy needs more capacity than the necessary information condition alone indicates (\Cref{tab:review-capacity}).

The shared channel spends $1/m$ reviews per step on anchoring and uses the remaining capacity for binary item decisions, where perfect review reveals a label and supplies $H(p)$ nats with $\omega=1$. The analytical screen combines a sufficient worst-case excursion bound with the necessary information condition in \Cref{eq:throughput}, so passing it does not establish that a policy can meet both targets. \Cref{tab:review-capacity} compares the first capacity that passes this screen with the first that meets the excursion target and expected log-growth condition under the tested policy.

For each integer cadence $m\in\{1,\ldots,60\}$, the policy reviews items independently with probability $\phi=\min\{1,(f_h-1/m)_+,(1-\bar r)/p\}$. It keeps reviewed good items and rejects reviewed bad items. When $\phi<1$, it keeps each unreviewed item with probability $\kappa=[\bar r-\phi(1-p)]/(1-\phi)$, provided this lies in $[0,1]$. This gives expected retention $\bar r$ and kept-corruption rate $(1-\phi)\kappa p$. We assess feasibility from that rate's expected log growth rather than a noisy finite-sample sign test, then check measured log growth in separate 20,000-item simulations at the selected capacities.

\begin{table}[htbp]
\caption{Review capacity per step on a grid spaced by 0.05. The tested policy requires more capacity than the analytical screen indicates. Retention 0.95 violates the information-independent retention limit and is excluded from feasible allocations.}
\label{tab:review-capacity}
\small
\begin{tabular}{rrr}
\toprule
Retention target & Analytical screen & Tested policy \\
\midrule
0.75 & 0.15 & 0.20 \\
0.85 & 0.15 & 0.30 \\
0.90 & 0.20 & 0.40 \\
0.95 & Infeasible & Infeasible \\
\bottomrule
\end{tabular}
\end{table}

An operator should therefore check the retention target before buying more reviews: in this model, retaining 95\% is infeasible even with perfect labels. For feasible targets, the chosen review policy still needs testing at its allocated capacity because passing the information screen alone does not establish performance.

\begin{figure}[htbp]
\centering\includegraphics[width=\linewidth]{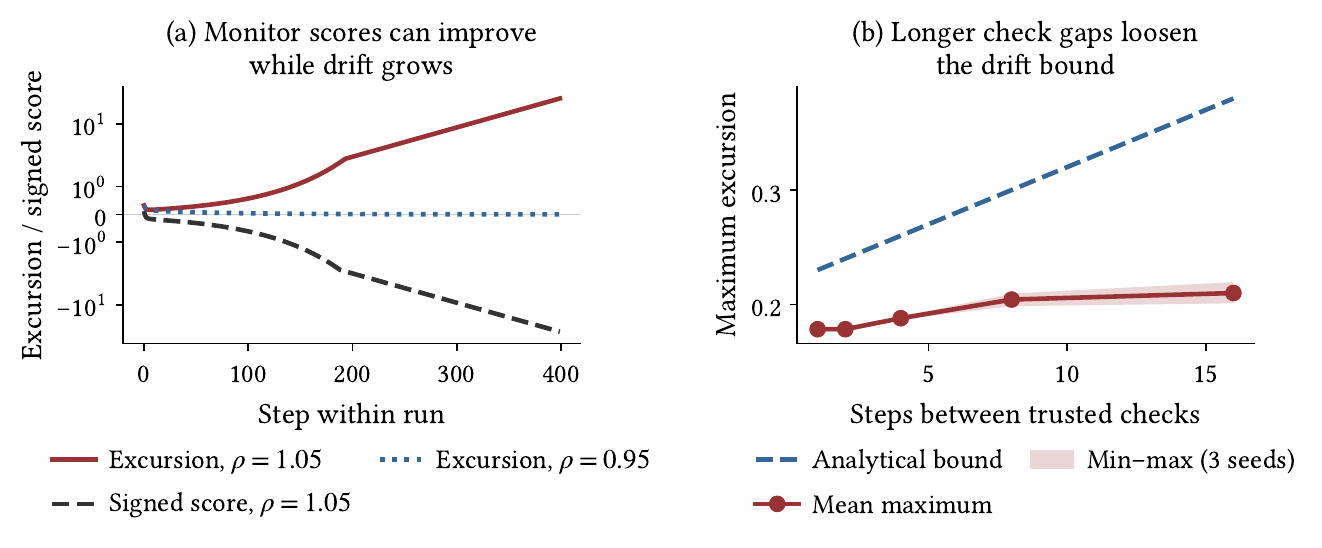}
\caption{Monitor scores can improve while departure from the charter grows (simulation). (a) The agent moves toward the monitor's reference, which also changes in response to the agent. With the stronger response, $\rho=1.05$, true excursion grows while the signed monitor score falls. With the weaker response, $\rho=0.95$, excursion converges toward zero. Negative scores indicate that the monitor prefers the agent's position to the charter reference, not negative excursion. The vertical axis is linear near zero and logarithmic farther away. (b) A separate model satisfies the bounded-error assumptions and remains below the analytical bound at every tested check interval. Points show mean maximum excursion over three seeds, and shading shows their observed minimum--maximum range, not a confidence interval. This second model does not establish a remedy for the failure in panel (a).}
\Description{Panel (a) plots deterministic trajectories on a symmetric logarithmic scale. Above the coupling threshold, true excursion rises while the signed monitor score falls below zero. Below the threshold, excursion approaches zero. Panel (b) plots the analytical bound and three-seed mean maximum excursion against steps between trusted checks. The shaded seed range remains below the bound.}
\label{fig:coupling}
\end{figure}

\paragraph{Monitor feedback and calibration settings}
Figure~\ref{fig:coupling}(a) uses the zero-bias model in \Cref{app:result-2}, with $\rho\in\{0.95,1.05\}$. Panel (b) summarizes trajectory maxima over three seeds at each check interval $m\in\{1,2,4,8,16\}$, using the anchored-residual settings listed above. In Figure~\ref{fig:coupling}(a), the agent moves toward the monitor's reference, while the reference itself changes in response to the agent. When the monitor responds more strongly to the agent, at $\rho=1.05$, true excursion grows even as the signed monitor score falls. This score compares the agent's position with the charter reference under the monitor's current judgment. A negative score means that the monitor prefers the agent's position, not that excursion is negative. Converting negative scores to zero would therefore leave the alarm silent. By contrast, at $\rho=0.95$, excursion converges toward zero. These trajectories come from a linear model, not observations of a language-model judge.

Figure~\ref{fig:coupling}(b) uses a separate model that satisfies the assumptions of \Cref{eq:anchored}. All three trajectories at each of five tested check intervals remain below the analytical bound. Longer intervals allow more error to accumulate between checks, increasing the bound. This result does not establish that periodic calibration repairs the unbounded feedback failure in panel (a).

\paragraph{Recovery and time available after updates}
The recovery simulation attains $\nrec(J)$ at all six tested jump sizes because it uses equality dynamics, checking the recurrence rather than establishing a minimum recovery time for every controller. By contrast, the repeated-update simulation compares a staged policy with one that continues operating without reassessment (\Cref{fig:app-dwell}). At each update, the simulator adds $J=2$, samples $a$ uniformly from $[0.85,0.95]$, and samples $|b|$ uniformly from $[0.8,1.2]$ with equal probability for either sign. Step disturbances are uniform on $[-0.05,0.05]$. The staged policy applies fallback until $y\le0.8$, then uses two opposite-sign probes to estimate the steering direction before testing whether an intervention satisfies the robust barrier condition. It operates only when that test passes, otherwise returning to fallback, whereas the comparison policy always operates and retains its initial positive-sign assumption. The plotted eight-step reference allowance combines six steps for nominal recovery with two for probes, but actual staging duration depends on the current score. For an update interval $\tau$, the reference operating fraction is $\max\{0,1-8/\tau\}$.

\section{Oversight duties and operational questions}
\label{app:governance}
For high-risk systems, the EU AI Act assigns human-oversight duties to providers and deployers and post-market monitoring to providers \cite[Arts.~14, 26, 72]{euaiact2024}. Meanwhile, the NIST AI RMF Playbook addresses third-party risk, monitoring, and recovery \cite{nistairmfplaybook}. \Cref{tab:regulation} connects these duties and practices to operational questions, and links them to the relevant limitations raised by our work.

\begin{table}[htbp]
\caption{Operational questions for implementing oversight duties and practices. Legal applicability and compliance require separate assessment.}
\label{tab:regulation}
\small
\begin{tabular}{p{0.28\linewidth}p{0.34\linewidth}p{0.29\linewidth}}
\toprule
Duty or practice & Operational question & Relevant limit \\
\midrule
EU AI Act Art.~14(4)(e): interruption & Can the operator stop pending actions and verify the resulting state? & Prior effects may require separate recovery \\
EU AI Act Art.~72: post-market monitoring & Do evaluations cover accumulated memory and configuration changes? & Inherited drift requires baseline evidence \\
EU AI Act Art.~43(4): substantial modification & What must be reassessed after this change, and can the chosen procedure finish before further changes? & The maintenance budget is policy-specific; legal triggers require separate analysis \\
NIST AI RMF MANAGE 3.1--3.2: third-party risk & Can model changes be detected, assessed, deferred, or reversed? & A vendor update may invalidate earlier calibration \\
NIST Playbook MANAGE 4.1: monitoring and recovery & Are exposure, residual harm, and intervention capacity measured separately? & Exposure and repair are distinct outcomes \\
\bottomrule
\end{tabular}
\end{table}
\end{document}